\documentclass[11 pt]{article}
\usepackage[margin=1in]{geometry}
\usepackage{bm}
\usepackage{titlesec}
\usepackage{pseudocode}
\usepackage{array,booktabs}
\usepackage{epsfig}
\usepackage{lscape}
\usepackage{graphicx}
\usepackage{setspace}
\usepackage{mathtools}
\usepackage{amsmath}
\usepackage{blkarray}
\usepackage{booktabs,tabularx}
\usepackage{threeparttable}
\usepackage{subcaption}
\usepackage{graphicx}
\usepackage{multirow}
\usepackage{indentfirst}
\usepackage{float}
\usepackage{amssymb}
\usepackage{natbib}
\usepackage{color}
\usepackage{xcolor}
\usepackage[justification=centering]{caption}
\usepackage{setspace}
\usepackage{hvfloat}
\usepackage{graphicx,rotating,booktabs}
\usepackage{algorithm}
\usepackage[noend]{algpseudocode}
\usepackage{afterpage}
\usepackage{pdflscape}

\usepackage{listings}
\usepackage{pythonhighlight}

\usepackage{pslatex}
\usepackage[T1]{fontenc}
\usepackage[utf8]{inputenc}
\usepackage{siunitx,booktabs}
\usepackage{array}
\usepackage{hyperref}
\hypersetup{
	pdfborder = {0 0 0}
}
\definecolor{darkblue}{rgb}{0,0,.4}
\hypersetup{colorlinks=true, breaklinks=true, citecolor=darkblue, linkcolor=darkblue, menucolor=darkblue, urlcolor=darkblue}

\usepackage[noabbrev]{cleveref}

\newcolumntype{L}[1]{>{\raggedright\let\newline\\\arraybackslash\hspace{0pt}}m{#1}}
\newcolumntype{C}[1]{>{\centering\let\newline\\\arraybackslash\hspace{0pt}}m{#1}}
\newcolumntype{R}[1]{>{\raggedleft\let\newline\\\arraybackslash\hspace{0pt}}m{#1}}

\titleformat{\paragraph}
{\normalfont\normalsize\bfseries}{\theparagraph}{1em}{}
\titlespacing*{\paragraph}
{0pt}{3.25ex plus 1ex minus .2ex}{1.5ex plus .2ex}

\title{From Live to Recording: Consumer Demand and Response to Price Across the Livestreaming Lifecycle}

\author{Ziwei Cong, Jia Liu, Puneet Manchanda%
	\thanks{Ziwei Cong is an Assistant Professor of Marketing, Georgetown University. Email: \href{mailto:Ziwei.Cong@georgetown.edu}{Ziwei.Cong@georgetown.edu}. Jia Liu is an Associate Professor of Marketing, Hong Kong University of Science and Technology. Email: \href{mailto:jialiu@ust.hk}{jialiu@ust.hk}. Puneet Manchanda is Isadore and Leon Winkelman Professor, Stephen M. Ross School of Business, University of Michigan. Email: \href{mailto:pmanchan@umich.edu}{pmanchan@umich.edu}. The paper is part of the first author's dissertation.}
}

\begin{document}
	
% spacing and title
\maketitle
\doublespacing
\thispagestyle{empty}

\begin{abstract}
\tiny \renewcommand{\baselinestretch}{1.1} \normalsize
\noindent
		
\noindent Livestreaming has evolved into a thriving industry where creators can directly monetize and engage with their audiences and followers. In practice, creators and platforms typically concentrate their marketing efforts on the period leading up to the livestream. However, livestreaming events naturally transition into recorded formats once the event concludes, creating potential ``residual'' opportunities for monetization. This study systematically examines consumer demand for live events throughout the entire livestream life-cycle, using data from a large livestreaming platform that allows consumers to purchase the recorded version of a paid live event after the livestream ends. We find that the demand is surprisingly more price-sensitive during the pre-livestream period compared to the post-period. This is partly driven by two mechanisms: consumer self-selection (infrequent consumers who may have missed the live events exhibit a higher willingness to pay for recorded versions) and quality uncertainty (consumers face higher uncertainty in event quality during the pre-period than in the post-period). Our findings generate implications for the pricing and targeting strategies in livestreaming markets.
		
\
		
\end{abstract}
	
\newpage
\setcounter{page}{1}
\pagenumbering{arabic}

\section{Introduction}\label{sec:intro}
	
\noindent In recent years, live streaming has grown into a widely popular media format. The global livestreaming economy was valued at approximately \$90 billion in 2023 and projected to reach \$400 billion by 2027 \citep{grandviewresearch}. To capitalize on this growth, platforms such as YouTube, Facebook, and Twitch have developed monetization strategies that allow creators to generate revenue through paid livestream events, subscriptions, and viewer donations. Such monetization strategies often focus on driving engagement and revenue during the pre-livestream and livestream periods, based on the intuitive belief that audience excitement predominantly resides in the live experience. However, livestreaming events naturally transition into recorded formats after the event is over, creating potential opportunities for sustained audience engagement and subsequent monetization. Overlooking the recorded versions may cause creators and platforms to miss out on important revenue sources and growth opportunities.
	
Despite the importance of recorded versions, the existing literature focuses primarily on live periods \citep{lu2021larger, lin2021express, simonov2023suspense, huang2025promotional}, leaving questions about the value of recorded versions largely unanswered. This paper aims to bridge this research gap by systematically analyzing consumer behavior across the entire lifecycle of livestreaming events. Our empirical study uses internal data from Zhihu, a large content platform in China. Zhihu started as a Q\&A community in 2011 and launched the Zhihu Live program in 2016, which allows users to host paid live events. Zhihu Live adopts the pay-per-view pricing model, with creators setting ticket prices for their own events (denoted as a ``Live'').\footnote{The pay-per-view model is also used by Facebook, Vimeo, and OnlyFans and is considered to be a ``direct'' pricing model, along with subscriptions, and pay-what-you-want or tipping \citep{lu2021larger,lin2021express}. Some platforms follow an ``indirect'' pricing model, i.e., revenue is generated via ad and/or product placement during the livestream.} Typically, a Live event is available for purchase 2-3 weeks before the livestream, and remains accessible at the same price after the event concludes. This setting is particularly suitable for our research objective. First, the pay-per-view model allows us to estimate the price elasticity of demand, a key metric to understand consumer valuation of the live and recorded content. Second, the automatic generation of recorded versions for each event facilitates a direct comparison of consumer responses to each format for the same event. Third, the price for each event is fixed over time, ensuring that any differences in consumer responses can be attributed to demand-side factors rather than price fluctuations.
	
Our empirical approach relies on cross-event data variation to estimate price elasticities. We address the endogeneity of pricing decisions using two complementary approaches. First, leveraging the platform's internal data, we estimate the price elasticity of demand using regression models that control for a comprehensive set of creator, event, and market factors that may influence both creator pricing decisions and sales. We assess the robustness of our results to omitted variable bias and model mis-specification using the bias-adjusted coefficient approach developed by \cite{oster2019unobservable} and Double Machine Learning (implemented with flexible nuisance functions). Second, we use the instrumental variable (IV) approach as an alternative identification strategy to infer the causal link between price and demand. Our IVs are based on the prices of similar events, which are identified using event embeddings (trained with Doc2Vec) derived from event descriptions and creator profiles. We discuss how our instruments satisfy the requirements of exclusiveness and exogeneity, using contextual information and supporting empirical evidence. Overall, we find consistent and robust results across different identification strategies and robustness checks.
	
Our findings show that the price elasticity is approximately $-1.2$ ($p < 0.01$) in the pre-livestream period, while it is $-0.8$ ($p < 0.01$) in the post-livestream period. This indicates that demand is more price-sensitive before the livestream, surprisingly suggesting a higher willingness to pay for the recorded version. We then explore the underlying factors driving this phenomenon. We find that the across-period gap in price elasticity is partly due to consumer self-selection. Given the limited time window before the livestream, consumers who notice and purchase these events early tend to be frequent users who have greater market knowledge and higher awareness of alternatives, making them more selective and sensitive to price. In contrast, post-period consumers tend to be infrequent users who have limited market familiarity and knowledge, resulting in relatively inelastic demand.
 
Nevertheless, we find that consumer self-selection does not fully account for the observed elasticity gap because the gap persists even after conditioning on consumer engagement levels. Thus, we further investigate an additional mechanism: quality uncertainty. Specifically, consumers face greater uncertainty regarding event quality in the pre-period, but this uncertainty diminishes in the post-period as quality is revealed. If this mechanism is indeed at play, the across-period elasticity gap should be smaller in cases where quality uncertainty is lower and the marginal benefit of post-event quality revelation is smaller. Our analysis confirms this hypothesis. We find smaller temporal gaps in price elasticity among creators with stronger reputational signals, in content categories that allow for more standardized quality assessments, and among consumers who are already familiar with the creators.

We further conduct policy simulations to examine how across-period differences in price elasticity affect revenue and pricing strategy. Specifically, we compare the current time-invariant pricing regime with a hypothetical period-specific pricing regime that allows creators to charge different prices in the pre- and post-event periods, under assumptions about creator sophistication and intertemporal demand independence. Under period-specific pricing, the optimal post-period price is 21.93\% higher than the pre-period price and 17.18\% higher than the optimal price under the time-invariant regime. On average, period-specific pricing increases revenue by 2.28\% relative to time-invariant pricing. This gain is disproportionately larger for less-established creators with weaker reputational signals, for whom post-period quality revelation substantially alleviates consumer uncertainty.

Our research makes important contributions to both academia and practice. First, we provide a comprehensive view of consumer engagement throughout the livestreaming lifecycle, spanning the pre-livestream stage and the subsequent recorded format. In doing so, we contribute to the literature on livestreaming \citep{lu2021larger, huang2025promotional}, consumer preferences for live versus recorded content \citep{vosgerau2006indeterminacy}, and advance selling \citep{xie2001electronic}. Second, our results reveal that while most platforms and creators currently concentrate their promotional efforts on the pre-livestream period, substantial revenue opportunities exist in the post-livestream phase. Platforms may benefit from rebalancing their marketing strategies to capitalize on the higher willingness to pay observed for the recorded version of livestreams. Finally, our findings imply the potential value of adopting dynamic pricing mechanisms in this market context. Given the rapid and global diffusion of livestreaming media, our research provides findings and implications that can be explored and potentially implemented in a wide variety of livestreaming platforms and creators.
	
The rest of the paper is organized as follows. We first review the relevant literature in \S \ref{sec:lit}. We then detail our empirical context, data, and descriptive analyses in \S \ref{sec:context}. Next, we clarify the identification and estimation of our model and show the main results in \S \ref{sec:main-iden}. We then discuss the potential underlying mechanisms for our results in \S \ref{sec:mechanism}, followed by the managerial implications for creators and platforms via policy simulations in \S \ref{sec:policy}. We conclude in \S \ref{sec:conclusion} with a discussion.

\section{Relevant Literature} \label{sec:lit}
	
\noindent Our paper relates to three streams of literature. The first is around the emerging phenomenon of livestreaming. Most existing studies focus on the ``live'' component, such as the drivers of viewership and engagement in livestreaming \citep{lu2021larger,lin2021express,simonov2023suspense} and the effectiveness of livestreaming promotion on consumer product usage and shopping behavior \citep{huang2025promotional}. We extend the research scope to the entire livestream lifecycle, from the live event itself to the subsequent recorded versions. This provides a holistic view of consumer engagement in the livestreaming markets and the distinct but complementary nature of live and recorded content.
	
The second is the understanding of consumer preferences for live versus recorded content. \cite{vosgerau2006indeterminacy} show that consumers prefer live television programming, such as football matches and lottery drawings, over tape-delayed broadcasts because of the indeterminacy of outcomes. \cite{rifkin2023preference} document that the experience of spontaneity explains consumer preferences for entertainment such as live court TV and online streaming services. While these studies emphasize the benefits of live content, our findings shed light on the conditions under which recorded content may be more valuable to consumers, thereby offering a comprehensive understanding of the trade-offs between live and recorded formats.

The third is the literature on advance selling \citep{dana1998advance, desiraju1999strategic}, which is common in many industries, such as travel, entertainment, and personal services. \cite{shugan2000advance} propose that consumers in the advance-selling period face uncertainty about their future consumption state (e.g., health, expected conflicts, mood). Building on this, \cite{xie2001electronic} further show that sellers can compensate consumers for state uncertainty by offering advance discounts. Our paper highlights a different type of uncertainty---quality uncertainty---which is particularly salient in the emerging creator economy, where sellers are often individuals with limited prior verifiable quality signals.\footnote{In our context, post-livestream recordings are available to consumers, mitigating state uncertainty and allowing us to focus on quality uncertainty.} Moreover, rather than focusing solely on advance selling, our study encompasses both the pre-livestream (advance selling) and post-livestream periods, shedding light on seller strategies across the entire livestream lifecycle.
	
\section{Context and Data}\label{sec:context}
	
\subsection{Zhihu Live}
	
\noindent Zhihu is the first and largest knowledge sharing platform in China, starting as a Q\&A community (similar to Quora.com) in 2011. The latest statistics show that Zhihu has over 43.1 million cumulative content creators and over 315.3 million queries and answers \citep{zhihurevenue2021}. Users on the Q\&A community can voluntarily seek and share information (such as knowledge, expertise, and customized solutions) in a variety of domains mainly through the posting of questions and answers. Users can also participate by evaluating answers through ``upvotes'' and ``downvotes'' and following other users organically. Each user has a profile page, which contains personal information provided by the user and a summary of the user's past content contribution activities. We provide a sample Zhihu interface in Web Appendix \ref{app:zhihu}.

\begin{figure}[!h]
	\centering
	\includegraphics[width=0.7\hsize]{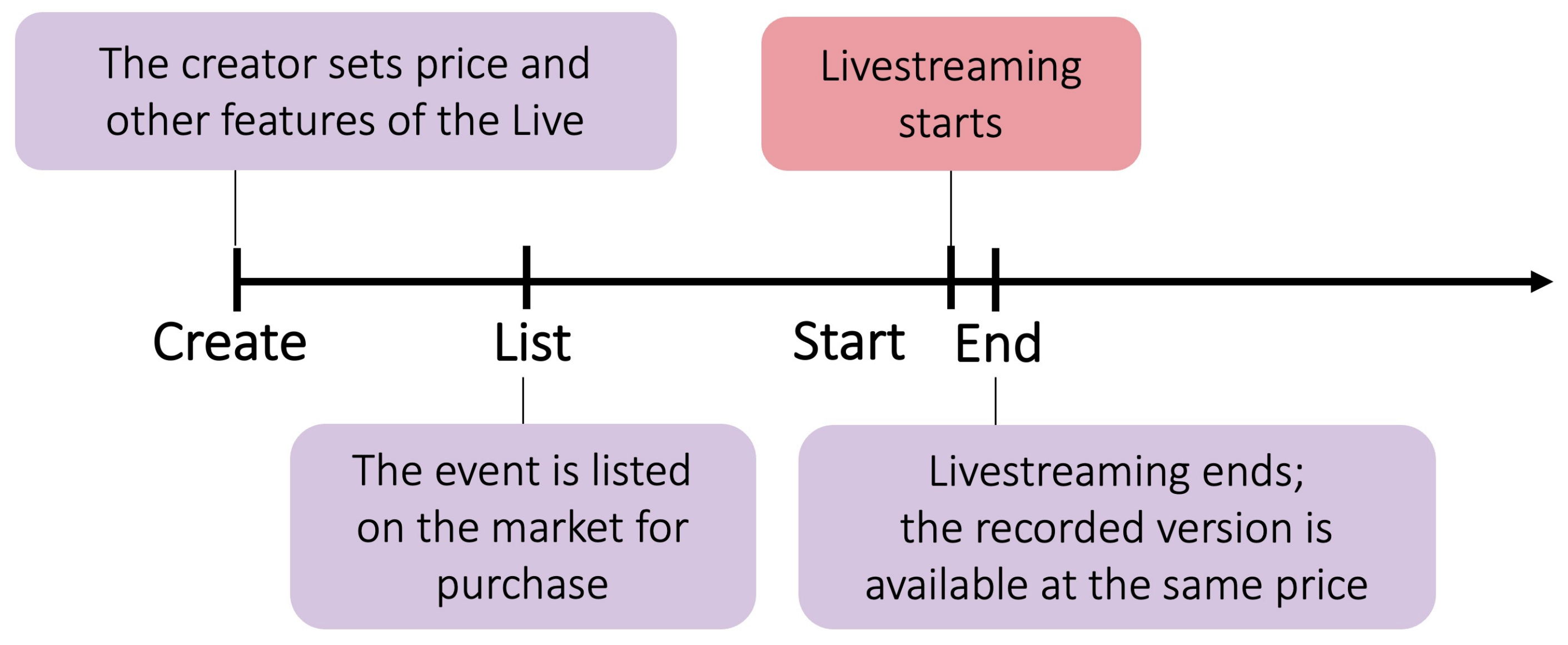}
	\caption{Typical Life-Cycle of a Live Event on Zhihu}\label{fig:live-life-cycle}
\end{figure}

Zhihu launched Zhihu Live on May 14, 2016, allowing users to monetize their expertise via a priced livestreaming event, the ``Live.'' By December 2020, Zhihu Live had nearly 10,000 Live events, 5,000 unique creators, and 6 million paying users \citep{zhihulive2020}. To become a Live creator, users need to provide Zhihu with basic personal information, such as real name, educational background, and proof of expertise in specific area(s). Upon approval, users are allowed to host their events. Figure \ref{fig:live-life-cycle} illustrates the life-cycle of a typical Live event. When creating a Live event, the creator decides on a few key features, including full price, topic, content, starting time, and seat limit. Creators have the freedom to set prices within a platform-specified range of 0 to 499.9 RMB.\footnote{To guide pricing, Zhihu provides a uniform price menu (e.g., 9.9, 19.9, 29.9, 39.9 RMB) for all events. Creators can choose a price from the menu or set a price outside the menu.} All event topics must fall under a set of 17 topical categories provided by Zhihu.\footnote{These are Architecture \& Design, Art History \& Appreciation, Business, Careers, Economics \& Finance, Education, Food \& Cuisine, Healthcare, Internet, Law, Lifestyle, Music \& Games \& Movies, Psychology, Reading \& Writing, Science \& Technology, Sports, and Travel.} The seat limit is the maximum number of users permitted to ask questions during the livestream, which is set at 500 for over 95\% of the events in our dataset.

% The cumulative number of sales is explicitly displayed on Live's webpage since its listing day.
Upon approval from Zhihu, the event is listed on the platform for purchase, typically 2-3 weeks before its starting time.\footnote{Details on how events are listed on the Zhihu Live homepage over time are provided in Section \ref{sec:iv-validity}.} During the livestream, the creator gives a live talk on the chosen topic by sending text, voice, and picture messages. The duration of the live session typically ranges between one and two hours. Moreover, the creator can interact with the audience by replying to their questions and comments. See an example in Web Appendix \ref{app:zhihu}. After the livestream concludes, consumers have the opportunity to provide feedback, by leaving a rating on a 5-point scale and open text comments. The Live recording is then made available for purchase at the same price, but without the opportunity to interact in real time with the creator.

Note that once creators set a price for their Live events, they cannot change the price after being listed for purchase. This prevents strategic price adjustments by creators in response to realized demand. The only possible within-event price variation comes from platform promotions, in which the platform offers discounts on a small subset of events (approximately 10\%). These discounted events are excluded from our study sample. In addition, the platform runs in-site banner advertising for a few events hosted by influential users or celebrities, which we control for in our study. These are the only marketing activities conducted by the platform during our study period. More details on the operation of the Zhihu Live website are provided in Section \ref{sec:iv-validity}.

% The only within-event price variations arise from platform promotions, in which the platform offers discounts on a small subset (approximately 10%) of Live events. These discounted events are excluded from our study sample.
	
\subsection{Data Description}\label{sec:data}
	
\noindent Our study uses two datasets from Zhihu. The first dataset contains transaction records of Live events from May 2016 to June 2017. We observe buyer ID, purchase time, unit price paid, and a rich set of characteristics describing the purchased Live event. We also observe advertising and promotions on events run by the platform, as well as ratings and reviews left by consumers for each event. The second dataset contains all users on Zhihu's Q\&A community along with their historical activities. These activities include registration time, content contribution, content subscription, social activities, and the number of different types of feedback that users receive for their contributed answers. As a result, we have access to detailed information on each creator and customer, as both creators and customers need to be registered on Zhihu to host/buy a Live event. Based on these comprehensive datasets, we construct a rich set of event and creator characteristics, shown in \Cref{tab:controls}, which will be used in our estimation.

Our empirical study focuses on events livestreamed between November 2016 and June 2017, a period when Zhihu Live market experienced no major structural or policy change. During this period, the platform promoted about 10\% of events hosted primarily by celebrities using a price discount, typically applied sometime between Live listing and streaming day. As noted earlier, we exclude these promoted events from our sample as they are not representative of the typical Live. The resulting sample contains 2,299 Lives hosted by 1,165 creators. Our outcome of interest is the sales volume of a Live event 30 days before and after the livestream (henceforth, pre-period and post-period), respectively. This window ensures comparability of post-livestream sales across events streamed in different weeks, while being comprehensive as it captures 85\% of total sales.
	
	\begin{table}[h!]
		\centering
		\small
		\begin{threeparttable}
			\caption{Control Variables in Main Specification}\label{tab:controls}
			\begin{tabular}{ll}  % Adjust the width of the columns as needed
				\hline
				\multicolumn{2}{c}{\textbf{Panel A: Event Characteristics}} \\
				\hline
				ExposureDuration & Number of days from listing to livestream\\
				SeatLimit & Max. number of users permitted to ask questions during the event \\
				Advertisement & Whether event $j$ is advertised by the platform\\
				Rating & Event $j$'s rating score \\
				Holiday & Whether event $j$ is streamed on a public holiday \\
				Event start week fixed effects & Event (livestream) week indicators \\
				Event start day-of-week fixed effects & Event (livestream) day-of-week indicators \\
				Event category fixed effects & Topical category indicators \\
				\hline
				\multicolumn{2}{c}{\textbf{Panel B: Creator Characteristics}} \\
				\hline
				\multicolumn{2}{l}{\textit{Characteristics within the QA Community and Additional Profile Information}} \\
				Follower/Followee & Cumulative number of followers and followees \\
				Answer & Cumulative number of answers contributed \\
				Upvote/Downvote/Thank/Unhelpful & Average number of upvote/downvote/thank/unhelpful per answer \\
				Badge & Whether creator $i$ receives Zhihu's ``Excellent Contributor'' badge \\
				TenureQA & Number of days since joining Zhihu \\
				Celebrity & Whether creator $i$ is an offline celebrity \\
				\hline
				\multicolumn{2}{l}{\textit{Characteristics within the Zhihu Live market}} \\
				NumPastEvent & Cumulative number of events hosted by creator $i$ \\
				AvgPastSales & Average sales of creator $i$'s past events \\
				AvgPastRating & Average rating of creator $i$'s past events \\
				TenureMarket & Number of days since creator $i$'s first event \\
				Invitation & Whether creator $i$ joins the Zhihu Live program by invitation \\		
				\hline
			\end{tabular}
			\begin{tablenotes}[para,flushleft]
				{\footnotesize Note: All creator characteristics are calculated at the start time of the Live event, except for \textit{Badge}, which is observed at the end of our study period. The variable \textit{AvgPastRating} is set to zero for each creator's first event after the launch of the Zhihu Live market.}
			\end{tablenotes}
		\end{threeparttable}
	\end{table}
		
\subsection{Summary Statistics}\label{sec:summary-stats}

\begin{figure*}[h!]
		\centering
		\begin{subfigure}[t]{0.45\textwidth}
			\centering
			\includegraphics[height=3in]{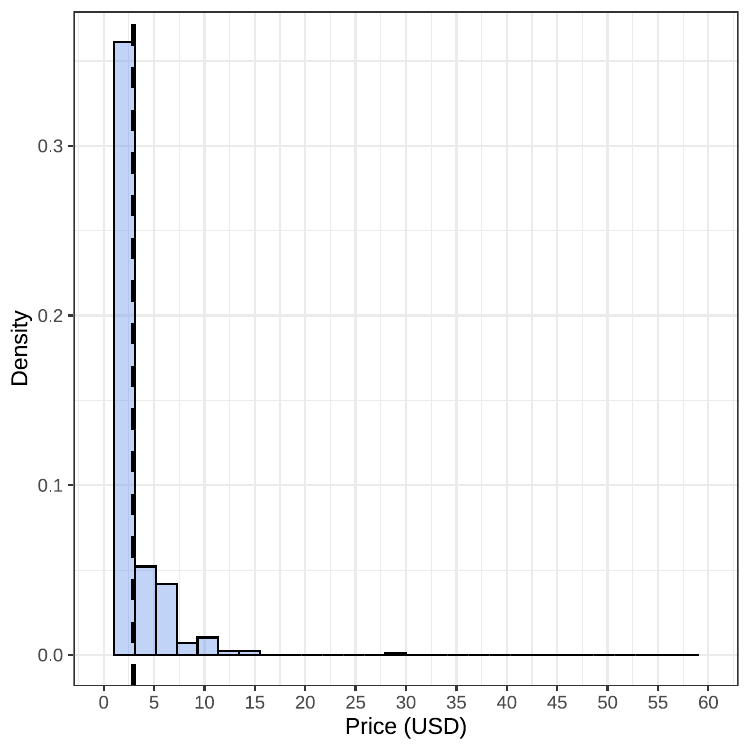}
			\caption{Across-event Distribution of Price}\label{fig:hist-price}
		\end{subfigure}%
		~
		\begin{subfigure}[t]{0.45\textwidth}
			\centering
			\includegraphics[height=3in]{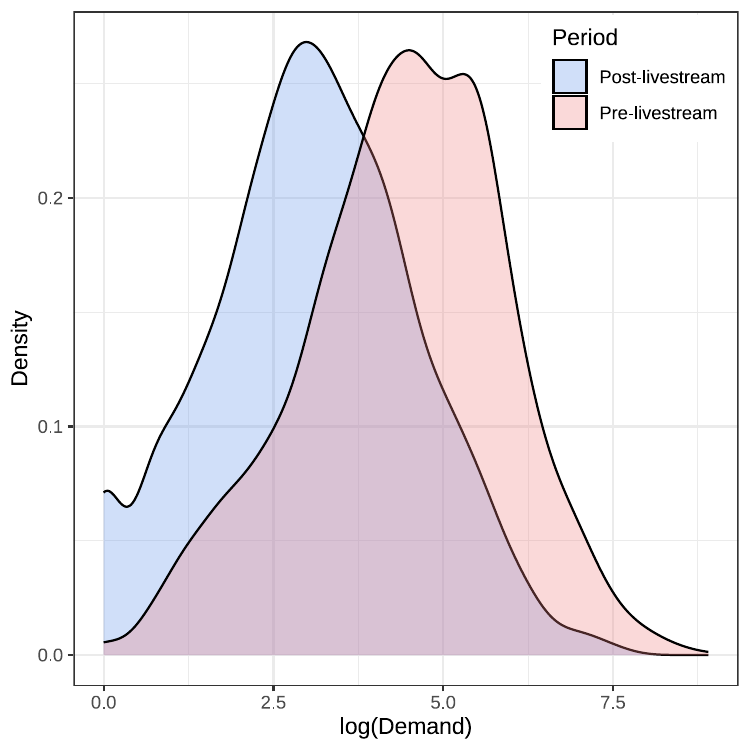}
			\caption{Across-event Distribution of Demand}\label{fig:hist-demand}
		\end{subfigure}
		\caption{Distribution of Price and Demand across Live Events}\label{fig:hist}
		\begin{minipage}{0.9\textwidth} % choose width suitably
			{\footnotesize Note: The dotted line (left panel) represents the average price (in USD) across all events.}
		\end{minipage}
	\end{figure*}

\noindent \Cref{tab:stat_price_demand} summarizes the key event characteristics. On average, an event is priced at around 3 USD (21 RMB), and has 218 tickets sold in the pre-livestream period and 65 in the post-period. Note that although post-period demand is smaller, it still accounts for 23\% of total sales, suggesting that post-livestream sales constitute an important revenue source for the platform and creators. \Cref{fig:hist} further presents the distributions of prices and (log-transformed) aggregated pre- and post-period demand, showing substantial heterogeneity across events. Notably, despite the lower average sales in the post-period, sales in the two periods exhibit similarly rich variations across events.

	\begin{table}[!h]
		\small
		\renewcommand{\arraystretch}{1.2}
		\addtolength{\tabcolsep}{0pt}
		\centering
		\begin{threeparttable}
			\caption{Summary Statistics of Demand and Price across Live Events}\label{tab:stat_price_demand}
			\begin{tabular}{l l r r r r r r r }
				\hline
				& & & & \multicolumn{5}{c}{Correlation} \\
				\cline{5-9}
				& Variable & Mean & Std. & V1 & V2 & V3 & V4 & V5 \\
				\hline
				V1 & Price (in USD) & 2.97 & 2.82 & 1 &  &   & & \\
				V2 & Pre-period sales & 217.54 & 424.72 & \textit{-0.157} & 1 & & & \\
				V3 & Post-period sales & 64.77 & 146.86 & \textit{-0.123} & \textit{0.743} & 1 & & \\
				V4 & ExposureDuration & 17.56 & 13.39 & \textit{0.154} & 0.010 & -0.007 & 1 & \\
				V5 & Advertisement & .04 & 0.24 & \textit{-0.058} & \textit{0.237} & \textit{0.097} & \textit{-0.050} & 1 \\
				\hline
			\end{tabular}
			\begin{tablenotes}[para,flushleft]
				{\footnotesize Note: The correlations in italics are statistically significant at $p<0.05$. }
			\end{tablenotes}
		\end{threeparttable}
	\end{table}

The correlation between price and demand, as shown in \Cref{tab:stat_price_demand}, is $-0.157$ and $-0.123$ for the pre- and post-period, respectively.\footnote{These results are based on the original scale of price and demand. When using log-transformed values (as in our main causal analysis), the correlations are $-0.300$ and $-0.212$, respectively.} These suggest an insensitive demand response to price. However, this result does not account for confounders (e.g., event or creator characteristics). As we will show subsequently, the effect of price on demand is more pronounced and also different across periods after controlling for confounders.

% The price distribution exhibits visible mass points, reflecting creators’ tendency to select from the platform’s recommended price menu, which includes commonly used price levels such as 9.9, 19.9, 29.9, and 39.9 RMB.

% \footnote{Twenty-one events in our sample are priced below the platform’s acceptable lower bound of 9.9 RMB (approximately 1.4 USD), with the lowest price set at 0.99 RMB (0.14 USD). These exceptions are typically tied to special occasions (e.g., Lunar New Year’s Eve, Valentine’s Day) or involve events hosted by highly prominent celebrities. Given their rarity, these outliers are unlikely to materially affect our estimation results.}
	
	\begin{table}[!h]
		\small
		\renewcommand{\arraystretch}{1.2}
		\addtolength{\tabcolsep}{0pt}
		\centering
		\begin{threeparttable}
			\caption{Creator Summary Statistics}\label{tab:statistics_speaker}
			\begin{tabular}{ll rrrr }
				\hline
				&   Variable  &  Mean & Std. & Min & Max  \\
				\hline
				&Follower                      & 23,117.19      & 62,893.72     &     0     &   1,278,507   \\
				&Followee                      & 156.01         & 337.09        &     0     &   5,379      \\
				&Answer                        & 225.58         & 457.75        &     0     &   4,792      \\
				&Upvote                        & 156.93         & 442.98        &     0     &   11,029     \\
				&Downvote                      & 4.81           & 11.40         &     0     &   208       \\
				&Thank                         & 48.80          & 1,483.89      &     0     &   5,443     \\
				&Unhelpful                     &  5.82          & 15.25         &     0     &   345       \\
				&Badge                         & 0.18           & 0.38          &     0     &   1         \\
				&TenureQA                      & 752.82         & 603.19        &     0     &   1,970     \\
				&Celebrity                     & 0.02           & 0.15          &     0     &   1        \\
				&NumPastEvent                  & 2.23           & 3.26          &     0     &   25       \\
				&AvgPastSales                  & 571.01         & 1,357.48      &     0     &   22,282   \\
				&AvgPastRating                 & 2.33           & 2.12          &     0     &   5        \\
				&TenureMarket                  & 61.21          & 78.92         &     0     &  388       \\
				&Invitation                    & 0.12           & 0.33          &     0     &   1        \\
				\hline
			\end{tabular}
			\begin{tablenotes}[para,flushleft]
				{\footnotesize Note: The number of observations is 2,299. Creator characteristics are calculated at the start time of their paid events, and these characteristics may exhibit slight variation over time.}
			\end{tablenotes}
		\end{threeparttable}
	\end{table}

Table \ref{tab:statistics_speaker} reports summary statistics of creator characteristics by the start time of each Live event. We find considerable variation in market experience, activity level, and reputation across creators, suggesting that our sample contains a diverse range of creators. Notably, 60\% (700) of all creators hosted only one event, 18\% (213) hosted two events, and 22\% (252) hosted three events and more. This suggests that generally, creators lack experience in selling and pricing paid content. Consequently, creators' pricing decisions may be somewhat arbitrary. To test this conjecture, we regress log-transformed price on all standardized creator and event characteristics defined in \Cref{tab:controls}. The estimation results are reported in \Cref{tab:ols_price}. We find that these characteristics collectively explain only 18.7\% of the total price variation. This highlights that a significant portion of price variations is driven by idiosyncratic factors beyond the systematic differences associated with event and creator characteristics. We discuss subsequently how such idiosyncratic variation may favor our causal identification. Another possible interpretation of the low R-sq is that there exist unobservable factors not captured in our data, but we provide empirical evidence in \Cref{sec:robu} that the impact of unobservables is minimal in our study.

	\begin{table}[!h]
		\footnotesize
		\renewcommand{\arraystretch}{1}
		\addtolength{\tabcolsep}{0pt}
		\centering
		\begin{threeparttable}
			\caption{Price Regression}\label{tab:ols_price}
			\begin{tabular}{ll *{1}{S[table-format=2.3, table-space-text-post=*********]}}
				\hline                    
				&  {Dependent Variable}  & {$Ln(Price)$} \\
				\hline
				\multicolumn{3}{l}{\textit{Event Characteristics}} \\
				&{ExposureDuration}                                & 0.061$^{***}$ \; (0.012)                   \\
				&{SeatLimit}                                       & 0.001 \; (0.010)                           \\
				&{Advertisement}                                   & -0.171$^{***}$ \; (0.039)                  \\
				&{Rating}                                          & -0.014 \; (0.013)                          \\
				&{Holiday}                                         & -0.084$^{*}$ \; (0.045)                    \\
				
				\multicolumn{3}{l}{\textit{Creator Profile in QA Community}} \\
				&{Follower}                                         & 0.063$^{***}$ \; (0.015)                  \\
				&{Followee}                                         & -0.003 \; (0.010)                         \\
				&{Answer}                                           & -0.003 \; (0.011)                         \\
				&{Upvote}                                           & -0.051 \; (0.035)                         \\
				&{Downvote}                                         & 0.028 \; (0.022)                          \\
				&{Thank}                                            & 0.048 \; (0.038)                          \\
				&{Unhelpful}                                        & -0.030 \; (0.044)                         \\
				&{Badge}                                            & 0.018 \; (0.029)                          \\
				&{TenureQA}                                         & -0.008 \; (0.011)                         \\		    
				&{Celebrity}                                        & 0.306$^{***}$ \; (0.067)                  \\
				
				\multicolumn{3}{l}{\textit{Creator Profile in Zhihu Live Market}} \\
				&{NumPastEvent}                                     & 0.102$^{***}$ \; (0.015)                  \\
				&{AvgPastSales}                                     & -0.015 \; (0.019)                         \\
				&{AvgPastRating}                                    & 0.044$^{***}$ \; (0.012)                  \\
				&{TenureMarket}                                     & -0.062$^{***}$ \; (0.018)                 \\
				&{Invitation}                                       & 0.200$^{***}$ \; (0.045)                  \\				
				
				\hline
                &Category fixed effects                             & {Yes}                                    \\
				&Time fixed effects                                 & {Yes}                                    \\
				\hline 
				& R$^2$                                             & {0.187}                                  \\
                & Num. obs.                                         & {2,299}                                  \\
				\hline
			\end{tabular}
			\begin{tablenotes}[para,flushleft]
				{\footnotesize Note: Time fixed effects refer to event start week and day-of-week fixed effects. We standardize all continuous covariates to ensure comparability of coefficients. $^{***}p<0.01$, $^{**}p<0.05$, $^*p<0.1$.}
			\end{tablenotes}
		\end{threeparttable}
	\end{table}

\section{Identification Strategies and Main Results}\label{sec:main-iden}

\noindent We measure the demand response to price by leveraging cross-event data variations. To address the endogeneity issues, we adopt two alternative identification strategies. First, taking advantage of the platform's internal data, we use regression methods to estimate the effect of price on demand by controlling for a rich set of observed creator, event, and market factors that may influence both creator pricing decisions and sales. We assess the robustness of our results to omitted variable bias and model mis-specification using the bias-adjusted method proposed by \cite{oster2019unobservable} and the Double Machine Learning (DML) method by \cite{chernozhukov2018double}. Second, we use the IV approach to infer the causal link between price and demand. Our instruments are based on the prices of similar events identified using event embeddings derived from event descriptions and creator profiles. In this section, we introduce each identification strategy and its corresponding estimation results. As we show below, the results are consistent across the different identification strategies.
	
\subsection{Model Specification}

\noindent To estimate and compare the price elasticity of demand for Live events in the pre- and post-periods, we specify the (log-transformed) aggregated sales of Live event $j$ during period $s \in \{Pre, Post\}$ as:
\begin{align}\label{eq:main}
Ln(Demand_{j,s}) = \alpha + \beta_{1} Post_{s} + \beta_{2} Ln(Price_{j}) + \beta_{3}Post_{s} \times Ln(Price_{j}) + \gamma X_{j_{it}} + \epsilon_{j,s},
\end{align}
where $Post_{s}$ is a period indicator equal to one for post-period and zero for pre-period, and $Price_{j}$ is the price set by creator $i$ at the time of event creation.\footnote{We apply the transformation $ln(x+1)$ to demand and price to accommodate zero demand observations.} The coefficient $\beta_{2}$ captures the price elasticity of demand in the pre-period, and $\beta_{3}$ captures the change in price elasticity in the post-period. Thus, the price elasticity in the post-period is $\beta_{2} + \beta_{3}$. The vector $X_{j_{it}}$ controls for a rich set of creator and event characteristics as listed in \Cref{tab:controls}. Creator characteristics include creator past activities and performances in the Zhihu Live market and on the Q\&A platform up to the event start time. Event characteristics include event topical category, whether the event is advertised by the platform, seat limit, rating score, etc. In addition, we control for event start time, using event start week and day of week fixed effects, which capture the general platform trends over time. We also allow these time fixed effects to interact with the $Post$ indicator, because some trends may have differential impacts on the pre- and post-periods. We standardize all continuous control variables for comparability across coefficients.\footnote{Note that an alternative model specification is to estimate demand separately for the pre- and post-periods. This specification enables the effects of covariates to vary across periods and permits the inclusion of different covariates in each period (e.g., incorporating the rating variable exclusively in the post-period). However, it does not allow a formal test of whether price elasticities differ significantly between periods. For this reason, we use Equation~\eqref{eq:main} as our primary specification and employ the separate regressions as robustness checks in Section \ref{sec:robu}.}
	
The main identification assumption underlying the above specification is that our rich control variables are sufficient to account for the potential confounders that influence both price and demand. This assumption is plausible in our context for three reasons. First, similar to most individual content creators, Zhihu creators tend to rely on their online (rather than offline) profiles and activities for pricing decisions.\footnote{It is important to note that 2\% of creators are offline celebrities, most of whom were invited by Zhihu to host paid events in the early stage of the Zhihu Live market as a promotional strategy. We use a ``celebrity'' dummy variable to control for creators' offline reputations.} This is particularly true given that event purchases are only open to registered Zhihu users, making creators' platform-based profiles the most important signals influencing demand. This also reduces creators' incentives for off-platform marketing, as such efforts are likely to be ineffective given the need to convert non-users into registered Zhihu users. Indeed, internal transaction data show that less than 1\% of sales originated from outside the platform during our study period, confirming the minimal impact of external marketing activities.
	
In addition, as discussed in \Cref{sec:summary-stats}, the majority of creators hosted only one or two Live events in our study period. With limited experience in selling paid events, creators' pricing decisions are likely influenced by idiosyncratic factors rather than strategic considerations. Many creators may have set prices arbitrarily or aligned their prices with the prices of other events available in the market. These idiosyncratic factors may introduce random components into pricing (especially when comparing among similar events and creators), which is also confirmed by the low R-sq value in \Cref{tab:ols_price}.
	
Lastly, our study is free from the issue of reverse causality, which is a common source of potential endogeneity in demand analysis. Typically, the relationship between demand and price could be bidirectional. For example, a seller could adjust the price in real-time in response to the observed demand patterns. This is unlikely in our context because creators cannot change the price once it has been set. Although the platform occasionally runs price promotions, as noted earlier, these apply to a very small number of events that have been excluded from our analysis.
	
\subsection{Main Results}

\begin{table}[!h]
\footnotesize
\renewcommand{\arraystretch}{1}
\addtolength{\tabcolsep}{0pt}
\centering
\begin{threeparttable}
\caption{OLS Estimation: Main Results}\label{tab:ols_main}
\begin{tabular}{ll *{3}{S[table-format=2.3, table-space-text-post=*********]}}
\hline
\multicolumn{2}{l}{Dependent Variable: $Ln(Demand)$} & \textit{No Controls} & & \textit{Full Controls}  \\
& & {(1)} & & {(2)} \\
\hline
\multicolumn{5}{l}{\textit{Key Independent Variables}} \\
&{\textbf{$Post$}}                    &  -1.682$^{***}$ \; (0.126) & & -2.384$^{***}$ \; (0.853)   \\
&{\textbf{$Ln(Price)$}}               &  -0.989$^{***}$ \; (0.067) & & -1.151$^{***}$ \; (0.058)   \\
&{\textbf{$Post \times Ln(Price)$}}   &  0.282$^{***}$ \; (0.094)  & &  0.329$^{***}$ \; (0.078)   \\
\multicolumn{5}{l}{\textit{Event Characteristics}} \\
&{ExposureDuration}                   &                             & & -0.004 \; (0.025)           \\
&{SeatLimit}                          &                             & & -0.063$^{***}$ \; (0.019)   \\
&{Advertisement}                      &                             & & 0.554$^{***}$ \; (0.079)    \\
&{Rating}                             &                             & & 0.469$^{***}$ \; (0.026)    \\
&{Holiday}                            &                             & & -0.391$^{***}$ \; (0.091)   \\
\multicolumn{5}{l}{\textit{Creator Profile in QA Community}} \\			
&{Follower}                           &                             & & 0.107$^{***}$ \; (0.030)    \\
&{Followee}                           &                             & & 0.024 \; (0.020)            \\
&{Answer}                             &                             & & 0.027 \; (0.022)            \\
&{Upvote}                             &                             & & 0.398$^{***}$ \; (0.070)    \\
&{Downvote}                           &                             & & 0.171$^{***}$ \; (0.044)    \\
&{Thank}                              &                             & & -0.054 \; (0.076)           \\
&{Unhelpful}                          &                             & & -0.340$^{***}$ \; (0.089)   \\
&{Badge}                              &                             & & -0.131$^{**}$ \; (0.057)    \\
&{TenureQA}                           &                             & & 0.084$^{***}$ \; (0.021)    \\		    
&{Celebrity}                          &                             & & 0.768$^{***}$ \; (0.134)    \\		
\multicolumn{5}{l}{\textit{Creator Profile in Zhihu Live Market}} \\
&{NumPastEvent}                       &                             & & -0.139$^{***}$ \; (0.030)   \\
&{AvgPastSales}                       &                             & & 0.304$^{***}$ \; (0.039)    \\
&{AvgPastRating}                      &                             & & 0.033 \; (0.023)            \\
&{TenureMarket}                       &                             & & -0.001 \; (0.035)           \\
&{Invitation}                         &                             & & 0.083 \; (0.090)            \\				
\hline
& Category fixed effects                        & {No}    & & {Yes}  \\
& Time fixed effects                            & {No}    & & {Yes}  \\
& $Post$ $\times$ Time fixed effects            & {No}    & & {Yes}  \\
\hline
& R$^2$                                         & {0.218} & & {0.498} \\
& Num. obs.                                     & {4,598} & & {4,598} \\
\hline
\end{tabular}
\begin{tablenotes}[para,flushleft]
{\footnotesize Note: Each event contributes one observation for the pre-period and one for the post-period, resulting in a total of $2,299 \times 2 = 4,598$ observations. Time fixed effects refer to event start week and day-of-week fixed effects. We standardize all continuous covariates to ensure comparability of coefficients. $^{***}p<0.01$, $^{**}p<0.05$, $^*p<0.1$.} 
\end{tablenotes}
\end{threeparttable}
\end{table}

\noindent Column (2) of \Cref{tab:ols_main} reports the estimation results based on Equation \eqref{eq:main}. The coefficient on $Ln(Price)$ and its interaction term is $-1.151$ ($p<0.01$) and $0.329$ ($p<0.01$), respectively. These estimates imply a price elasticity of $-1.151$ in the pre-period and $-1.151 + 0.329 = -0.822$ in the post-period. Their magnitudes are more than 16\% larger than those obtained from a simple linear regression without any controls (see Column (1) of \Cref{tab:ols_main}). This comparison confirms that omitting creator and event characteristics likely results in underestimated price coefficients.\footnote{In both regressions, the coefficient on $Post$ is statistically significant and negative. It reflects the difference in baseline demand between the pre- and post-periods at a hypothetical price of zero---a value not observed in our data. As this coefficient lacks practical interpretability, we do not discuss it further.} Importantly, the post-period effect is nearly 30\% less elastic than that during the pre-period ($p<0.01$), suggesting that recorded versions have substantial residual value. This is economically meaningful, as post-period demand accounts for 23\% of total sales. We explore the underlying drivers of this residual value in Section \ref{sec:mechanism}.

Another component of Live events is the live experience. If the live experience has dominant value, we would observe lower price sensitivity in the pre-period. Thus, our results suggest that the live experience may be less appealing than commonly assumed.\footnote{\Cref{fig:ranking_effect} in Web Appendix \ref{app:mechanism} further shows that price elasticity remains flat during the weeks immediately preceding the livestream, indicating that anticipation of live interaction does not meaningfully influence demand. This provides additional evidence that the value of live interaction is limited in our setting. A detailed discussion of \Cref{fig:ranking_effect} is in \Cref{sec:alternative_explanation}.} This finding contrasts the one in \cite{vosgerau2006indeterminacy}, which documents a preference for live television programming (e.g., sports events and lottery drawings) over delayed broadcasts. One possible explanation is that the content in our setting is primarily learning-oriented and does not involve uncertain outcomes, thereby making the value of real-time consumption less salient.

Notably, the estimated price elasticity is relatively small, particularly during the post-livestream period (ranging between $-1$ and 0). This raises the question: Why don't creators raise their prices? One possible reason is the presence of established norms (or regulations) for livestreaming events that exist in the market. To explore this, we compare the average price of Zhihu Live events to other similar paid events, including livestreams, videos, and audio content available on other platforms during our study period. The details are presented in Web Appendix \ref{app:comp-price}. We find that Zhihu's prices represent the upper bound for such events, suggesting limited scope for Zhihu creators to increase their prices. In addition, as demand is more price elastic in the pre-livestream period and Zhihu did not use dynamic pricing during our study period, creators may be concerned that higher prices could negatively impact pre-livestream sales. These suggest the potential benefits of implementing period-specific pricing, which we will discuss in \Cref{sec:policy}.
	
\subsection{Robustness Checks}\label{sec:robu}
	
\noindent While the platform's internal dataset allows us to closely approximate the information set available to creators at the time of pricing, there may exist unobserved factors that could bias the (causal) estimates. The direction of the bias is unclear. For example, if desirable unobserved creator characteristics are positively correlated with price, then estimation without regard to this correlation will result in a price coefficient that is biased downward in magnitude.\footnote{The reason is clear: since higher prices are associated with desirable features of the event and creator, consumers avoid the higher-priced products less than they would if the higher prices occurred without any compensating change in unobserved attributes. Essentially, the estimated price coefficient picks up both the price effect (which is negative) and the effect of the desirable unobserved attributes (which is positive), with the latter muting the former.} A similar bias, though in the opposite direction, occurs if unobserved marketing efforts consist of advertising or other promotions coupled with a lower price\citep{train2009discrete,bray2024observational}.\footnote{The higher demand for marketed events comes from both the lower price and the non-price advertising or promotions. The estimated price coefficient picks up both effects, which are greater than the impact of lower prices by themselves.} The latter is less likely given that our datasets comprehensively record advertising activities within the platform and that the impact of external advertising is minimal. To examine the sensitivity of our findings to omitted variable bias, we conduct three sets of analyses. We report a summary below, with detailed results in Web Appendix \ref{app:robu_ols}.

\subsubsection{Inclusion of Additional Control Variables}\label{sec:robu_controls}

\noindent While time fixed effects account for unobserved temporal shocks that may simultaneously influence price and demand, such trends may vary systematically across topics. To mitigate this concern, we include the interactions between category fixed effects and time fixed effects (i.e., event start week and day of week fixed effects) as additional controls in Equation \eqref{eq:main}. The results, shown in \Cref{tab:ols_add_control} of Web Appendix \ref{app:robu_ols}, remain unchanged relative to our main results.

The effects of covariates may also differ across the pre- and post-periods. While Equation~\eqref{eq:main} accounts for this by interacting $Post$ with time fixed effects, it assumes that other covariates have homogeneous effects across periods. To relax this, Column (2) of \Cref{tab:ols_add_control} interacts all covariates with $Post$. This more flexible specification yields results consistent with our main results. An equivalent approach is to estimate demand separately for each period.\footnote{This approach also enables the inclusion of different covariates in each period, which cannot be achieved in the pooled specification of Equation \eqref{eq:main}. In Column (5) of \Cref{tab:ols_separate}, we exclude the rating variable from the pre-period regression, as this information is only available after the livestream. The estimated pre-period price elasticity is robust to controlling for the rating or not.} Columns (3) and (4) in \Cref{tab:ols_separate} report the separate regressions, which estimate price elasticity identical to Column (2) of \Cref{tab:ols_add_control}.

\subsubsection{Bias-adjusted Coefficients}
	
\noindent We next conduct a sensitivity analysis following the approach proposed by \cite{oster2019unobservable}. This approach examines the robustness of treatment effect estimation by observing coefficient and R-sq movements after the inclusion of controls. For ease of interpretation, we use the aforementioned separate regression framework for this analysis and the remaining robustness checks. Consider the following regressions:
\begin{align}
        Ln(Demand) &= \alpha + \beta_{0} Ln(Price) + \epsilon, \label{eq:oster1} \\
	Ln(Demand) &= \alpha + \beta_{1} Ln(Price) + X_{o} + \epsilon, \label{eq:oster2} \\
	Ln(Demand) &= \alpha + \beta^{*} Ln(Price) + X_{o} + X_{u} + \epsilon, \label{eq:oster3}
\end{align}
where $X_{o} = {\Phi}x_{o}$ represents the impact of observables and $X_{u}$ represents the impact of unobservables. Thus, Equation \eqref{eq:oster1} is the baseline regression without controls. Equation \eqref{eq:oster2} controls for observables. Equation \eqref{eq:oster3} is a hypothetical regression that accounts for both observables and unobservables, which yields the bias-adjusted treatment effect $\beta^{*}$. \cite{oster2019unobservable} proposes an approximation of the bias-adjusted treatment effect:
\begin{align}
	& \beta^{*} = \beta_{1} - \delta \left[ \beta_{0} - \beta_{1} \right] \frac{R_{max}^{2} - R_{1}^{2}}{R_{1}^{2} - R_{0}^{2}},
\end{align}
where $R_{0}^{2}$, $R_{1}^{2}$, and $R_{max}^{2}$ are the $R$-sq values from Equation \eqref{eq:oster1} to \eqref{eq:oster3}; the variable $\delta$ is the relative degree of selection on unobservables to observables.\footnote{Specifically, define the proportional selection relationship as $ \delta \frac{\sigma_{o,price}}{\sigma_{o}^{2}} = \frac{\sigma_{u,price}}{\sigma_{u}^{2}} $, where $\sigma_{s,price} = cov(X_{s}, price)$, $\sigma_{s}^{2} = var(X_{s})$ for $s \in \{ o, u \}$. Thus, $\delta$ can be interpreted as the extent to which the bias from observed characteristics is informative about the bias from unobserved characteristics.} A value of $\delta = 1$ suggests that the unobservables are as important as the observables.
	
We take estimates for $\beta_{0}$ and $R_{0}^{2}$ from \Cref{tab:ols_separate} in Web Appendix \ref{app:robu_ols}, Column (1) and (2); and estimates for $\beta_{1}$ and $R_{1}^{2}$ from Column (3) and (4). Following \cite{oster2019unobservable}, we set $\delta=1$ and $R_{max}^{2}=1.3 \times R_{1}^{2}$. \Cref{tab:bias-adj-coef} summarizes those estimates and the derived bias-adjusted coefficients $\beta^{*}$. We find that the value of $\beta^{*}$ remains similar to our main results. The interval formed by $\beta^{*}$ and $\beta_{1}$ also excludes zero, further confirming the robustness of our results \citep{oster2019unobservable}.\footnote{The bias-adjusted treatment effect is slightly more negative than our main estimates. This indicates that, though negligible, the impact of unobservables likely includes desirable event/creator features that positively correlate with price, rather than unobserved marketing efforts coupled with lower prices. This observation supports our conjecture regarding the direction and sources of omitted variable bias, confirming our understanding of the data and context.}
	
\begin{table}[!h]
\centering
\caption{Bias-adjusted Coefficients and $R_{max}^2$ Values for Pre- and Post-periods}
\label{tab:bias-adj-coef}
\begin{tabular}{lcccccc} 
        \hline
	& \textbf{$\beta_{0}$} & \textbf{$\beta_{1}$} & \textbf{$R_{0}^{2}$} & \textbf{$R_{1}^{2}$} & \textbf{$R_{\text{max}}^{2}$} & \textbf{$\beta^{*}$} \\ 
	\hline
	Pre-period   & -0.989  & -1.177   & 0.090  & 0.461    & 0.599  & -1.270 \\ 
	Post-period  & -0.707  & -0.795   & 0.045  & 0.352    & 0.458  & -0.836 \\ 
	\hline
	\end{tabular}
\end{table}

% \footnote{One reason to favor $\delta = 1$ as an appropriate cutoff is that researchers typically focus their data collection efforts (or their choice of regression controls) on the controls they believe ex ante are the most important \citep{oster2019unobservable, angrist2010credibility}.}

A final sensitivity analysis can be conducted based on the value of $\delta$ required to nullify the results, given an assumed $R_{\text{max}}^{2}$ value \citep{oster2019unobservable}. A required value of $\delta > 1$ indicates that unobservables must be more important than observables to produce a treatment effect of zero, thereby implying robustness to omitted variable bias. In our setting, the required $\delta$ is 44.11 for the pre-period and 21.66 for the post-period, assuming $R_{\text{max}}^{2} = 1.3 \times R_{1}^{2}$. Thus, unobservables would need to be 44 (pre-period) and 22 (post-period) times as important as observables to explain away the effects. Such implausibly large $\delta$ further confirms the validity of our findings.\footnote{Given our research objective, we also examine the strength of unobservables necessary to overturn our substantive conclusions. That is, to render pre-period demand inelastic and post-period demand elastic. To this end, we compute the value of $\delta$ required to yield a price coefficient of $-1$ in each period. We find that the required $\delta$ is 2.43 in the pre-period and 3.76 in the post-period (again assuming $R_{\text{max}}^{2} = 1.3 \times R_{1}^{2}$). These values remain large, further reinforcing the robustness of our conclusions.}
		
\subsubsection{Alternative Functional Forms} 
	
\noindent Omitted variable bias may also arise from model mis-specification. To mitigate this concern, we employ DML \citep{chernozhukov2018double}, which is well-known for its robustness to model mis-specification and regularization bias \citep{ellickson2022estimating}. In our implementation (detailed in Web Appendix \ref{app:DML}), we alternately use polynomial Lasso and Random Forest to more flexibly estimate the relationships between confounders and price (demand). The polynomial Lasso captures potential nonlinear relationships by selectively including the quadratic terms of every continuous covariate, while the Random Forest model further relaxes parametric assumptions on functional form. We find highly similar estimates across models (see \Cref{fig:dml} in Web Appendix \ref{app:robu_ols}). This observation suggests that functional form mis-specification is unlikely to be a major concern in our study.
	
\subsection{Instrumental Variable Approach}
	
\noindent We now handle potential omitted variable bias using the IV approach. In our context, a valid IV should cause variation in creators' pricing decisions (the relevance condition), but not directly affect the sales of creators' Live events (the exclusion restriction condition), and not be correlated with the error term in the outcome model (the exogeneity condition). Inspired by Hausman-type instruments \citep{hausman1994valuation,nevo2001measuring} that prices from other markets are used as instruments for prices in the focal market, we consider the prices of events similar to the focal event. This IV strategy is similarly used in \cite{tian2024mega} to measure follower elasticity. For ease of exposition, we refer to our IV as ``similar event prices.''
	
\subsubsection{IV Construction}
	
\noindent We first use Doc2Vec \citep{le2014distributed} to generate event embeddings based on their titles, detailed descriptions, and creator bios for 4,095 events in our original datasets.\footnote{These texts were prepared by creators at the time of event creation and were visible to consumers when events became available for purchase. We use the complete set of events from our original dataset to retain the most comprehensive information for training purposes, though our causal analysis specifically focuses on 2,299 events that were hosted between November 2016 and June 2017. For text preprocessing, we use Jieba Fenci to segment the text into individual words, and remove words that appear fewer than two times to reduce noise.} Our learning objective is to maximize the likelihood of accurately predicting a word based on its document vector and the vectors of surrounding words. The model is trained on a random 80\% sample of the events, with the remaining 20\% reserved for validation. The final Doc2Vec model produces a 40-dimensional vector for each event.\footnote{As a robustness check, we also construct event embeddings using OpenAI’s pre-trained embedding models, which are trained on large-scale corpora and designed to generate general-purpose representations applicable across diverse tasks. The IV estimates based on these embeddings (see Web Appendix \ref{app:robu_iv}) yield results that are consistent with those obtained using Doc2Vec-based embeddings. This consistency suggests that our findings are not sensitive to the choice of embedding method.} To assess our model, we conduct a ``sanity check'' by inferring a new vector for each event in the training set and comparing it to the vector generated during training. While this is not a formal accuracy test, it provides insights into whether the model behaves consistently and reliably. Our results show that over 99\% of the inferred representations are most similar to themselves (their corresponding vectors during training), confirming the validity of our model.

\begin{table}[!h]
    \centering
    \renewcommand{\arraystretch}{1}
    \addtolength{\tabcolsep}{0pt}
	\begin{threeparttable}
		\caption{Sampled Focal Events and Their Similar Events}\label{tab:doc2vec-vali}
		\begin{tabular}{l  l}
			\hline
            \multicolumn{2}{l}{Event title: How to think about data like an economist} \\
  1. & Unveiling the non-performing assets of commercial banks \\
			2. & Production, screening, and use of data \\
			3. & Self-service data analysis \\
			\hline
            \multicolumn{2}{l}{Event title: Beginner's guide to the kitchen}  \\
            1. & Design your kitchen like design a lab \\
            2. & Guide to inspecting a fully renovated home \\
            3. & Buying kitchen supplies \\
			\hline
		\end{tabular}
		\begin{tablenotes}[para,flushleft]\vspace*{-0.1in}
			{\footnotesize Note: The similar events are listed in descending order of cosine similarity with the focal event.}
		\end{tablenotes}
	\end{threeparttable}
\end{table}
	
For a given focal event $j$, we compute the cosine similarity between event $j$'s representation vector $r_{j}$ and the representations of all other events that were streamed at least \textit{seven days} before the creation time of event $j$. The seven-day cutoff mitigates concerns about demand competition between similar events, which we detail next. Then, we identify the most-, second-, and third-closest events to each focal event $j$, and construct its instruments as the log-transformed prices of these events, denoted as $Z_{j} = \{z_{1}, z_{2}, z_{3}\}$.\footnote{We note that our results remain consistent when we instead use the average price of these events to construct a single instrumental variable (See Web Appendix \ref{app:robu_iv}).} On average, the cosine similarity between the focal event and its top-3 most similar events is 0.720, and $53.55\%$ of the top-3 most similar events share the same category as the focal event. For interested readers, \Cref{tab:doc2vec-vali} illustrates two sampled events and their top-3 similar events.

\subsubsection{IV Validity}\label{sec:iv-validity}
	
\noindent We now describe how this IV meets the relevance, exclusiveness, and exogeneity criteria. The relevance condition likely holds for at least two contextual reasons: (1) similar events share comparable characteristics with the focal event, and (2) creators typically set prices by referencing the prices of similar events. Empirically, we find supporting evidence that the weak instrument test \citep{stock2002testing} yields an F-statistic of 286.511, which exceeds the typical threshold of 10 for determining relevancy.\footnote{We further conduct a falsification test: if creators price their events by referencing similar events but not dissimilar ones or market-wide pricing patterns, then dissimilar event prices should not be correlated with the focal event price. To test for this, we identify the top-3 most dissimilar events to each focal event and regress the focal event price on their prices. The estimated coefficients are statistically insignificant ($p>0.1$) and the R-sq is 0.001, indicating that dissimilar event prices have little predictive power over the focal event price. This supports the interpretation and construction of our instrument.}

The exclusiveness and exogeneity conditions require that similar event prices do not affect the sales of a focal event, except through its correlation with the focal event's price, or in ways that we control for explicitly in our model. On Zhihu, the primary way through which users can discover and shop Live events is the main page of the Zhihu Live market, which contains a non-personalized list of all available Live events. Thus, one major threat to our instrument's validity is the possibility that consumers make direct price comparisons or substitutions between similar events. However, several features of the Zhihu Live market, discussed below, block this potential causal pathway and preserve the exogeneity restriction.\footnote{We conduct an overidentification test \citep{sargan1958estimation} to assess the exogeneity assumption. The test yields a statistic of $7.130$ with a $p$-value of $0.129$, failing to reject the null hypothesis that all instruments are valid. This result provides suggestive evidence supporting the exogeneity assumption. However, the test relies on the strong assumption of homoskedasticity. We therefore place greater emphasis on institutional details to substantiate the plausibility of the exogeneity and exclusion restriction conditions.}

\begin{figure}[!h]
	\centering
	\includegraphics[width=1\hsize]{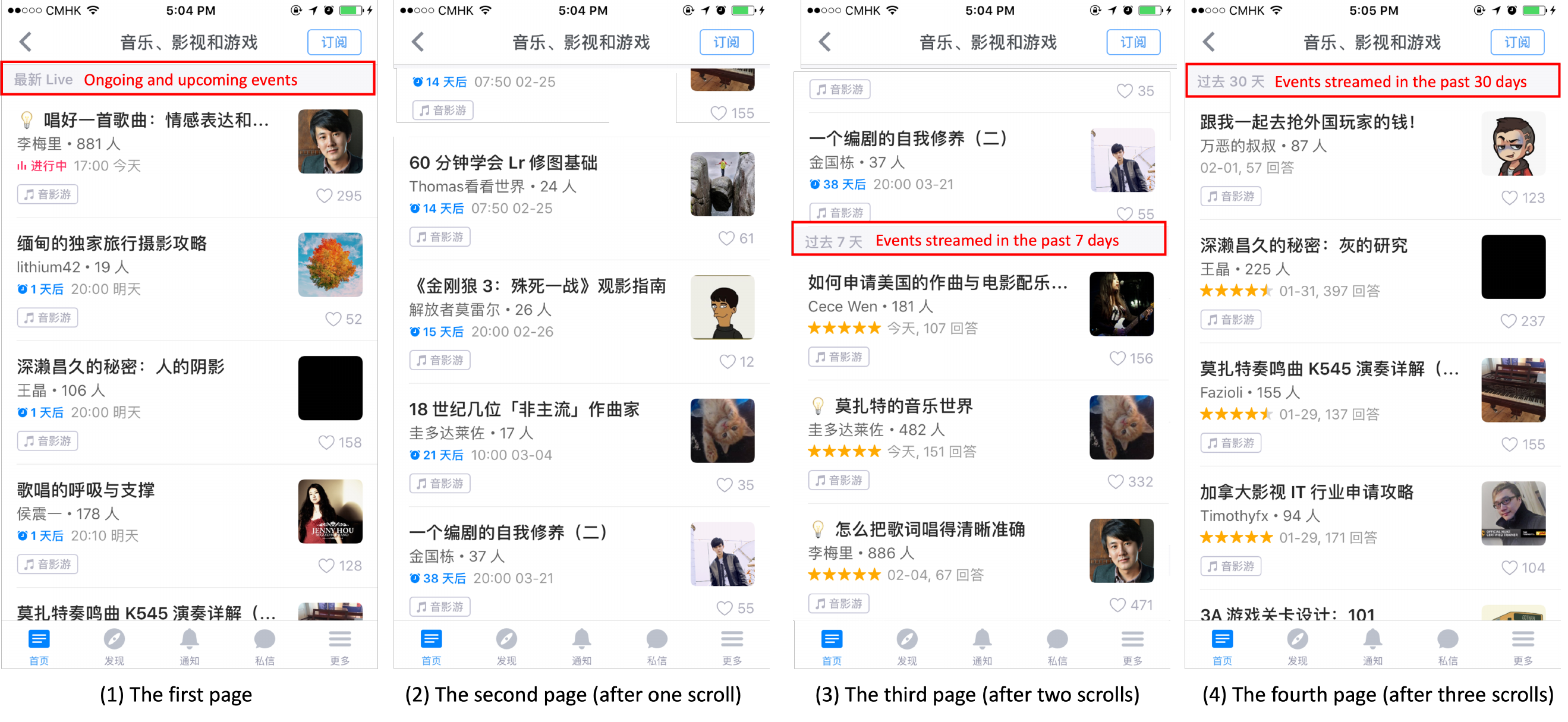}
	\caption{Layout of Zhihu Live Website}\label{fig:market-layout2}
	\begin{minipage}{0.9\textwidth} % choose width suitably
		{\footnotesize Note: These sample screenshots of the Zhihu Live marketplace (for the ``Music \& Games \& Movies'' category) were taken around April 2017. The layouts for other categories, as well as the overall marketplace layout without category filters, follow a similar structure.}
	\end{minipage}
\end{figure}

First, ongoing and upcoming events are featured prominently at the top of the page, while events from the past 7/30 days are listed further down, requiring extensive scrolling to access. See a sample main page in Figure \ref{fig:market-layout2}. This separation effectively isolates the markets for new and previous events, minimizing the potential for ``cross-border shopping'' or direct comparisons between similar events that are separated by a week or longer. This argument is consistent with industry statistics that over 80\% of clicks occur on the first page on digital platforms \citep{yu2024welfare,googleclick}. Our instrument construction explicitly leverages this layout by identifying similar events only from those streamed at least seven days before the focal event. Second, consumers must click into the event page to see prices, making direct price comparisons costly. Last, Zhihu did not provide a built-in search function for Zhihu Live events during our study period. Users could only filter events by broad categories, with no option to narrow results by personal interests, price, or event similarity.

A second channel through which similar event prices might directly affect focal event demand is shared popularity trends. If a particular type of content experiences a temporary surge in interest, both the focal event's sales and the prices of similar events, especially those scheduled in close temporal proximity, may rise simultaneously. Our instrument, which relies on similar events streamed at least seven days prior to the focal event, helps mitigate this concern. Our main specification controls for a rich set of event and creator characteristics, along with category and event start time fixed effects, which also help block this potential causal pathway. To further mitigate this concern, we include the event representation vector $r_{j}$ as additional controls for event content in our main specification. This inclusion does not change our results (see \Cref{tab:IV_additional_controls} Web Appendix \ref{app:robu_iv}).

\begin{table}[!h]
\footnotesize
\renewcommand{\arraystretch}{1}
\addtolength{\tabcolsep}{0pt}
\centering
\begin{threeparttable}
\caption{IV Estimation: Main Results}\label{tab:iv_main}
\begin{tabular}{ll *{1}{S[table-format=2.3, table-space-text-post=*********]}}
\hline   
\multicolumn{2}{l}{Dependent Variable: $Ln(Demand)$} & \textit{Full Controls} \\
\multicolumn{2}{l}{} & {(1)} \\
\hline
\multicolumn{3}{l}{\textit{Key Independent Variables}} \\
& \textbf{$Post$}                  & -2.565$^{***}$ \; (0.862)   \\
& \textbf{$Ln(Price)$}             & -1.270$^{***}$ \; (0.106)   \\
& \textbf{$Post \times Ln(Price)$} &  0.500$^{***}$ \; (0.138)   \\
\multicolumn{3}{l}{\textit{Event Characteristics}} \\
& ExposureDuration                 & -0.002 \; (0.025)           \\
& SeatLimit                        & -0.063$^{***}$ \; (0.019)   \\
& Advertisement                    &  0.548$^{***}$ \; (0.080)   \\
& Rating                           &  0.469$^{***}$ \; (0.026)   \\
& Holiday                          & -0.394$^{***}$ \; (0.091)   \\
\multicolumn{3}{l}{\textit{Creator Profile in QA Community}} \\			
& Follower                         &  0.109$^{***}$ \; (0.030)   \\
& Followee                         &  0.024 \; (0.020)           \\
& Answer                           &  0.026 \; (0.022)           \\
& Upvote                           &  0.396$^{***}$ \; (0.070)   \\
& Downvote                         &  0.172$^{***}$ \; (0.044)   \\
& Thank                            & -0.052 \; (0.076)           \\
& Unhelpful                        & -0.341$^{***}$ \; (0.089)   \\
& Badge                            & -0.130$^{**}$ \; (0.057)    \\
& TenureQA                         &  0.084$^{***}$ \; (0.021)   \\		    
& Celebrity                        &  0.778$^{***}$ \; (0.136)   \\		
\multicolumn{3}{l}{\textit{Creator Profile in Zhihu Live Market}} \\
& NumPastEvent                     & -0.135$^{***}$ \; (0.031)   \\
& AvgPastSales                     &  0.304$^{***}$ \; (0.039)   \\
& AvgPastRating                    &  0.035 \; (0.024)           \\
& TenureMarket                     & -0.003 \; (0.036)           \\
& Invitation                       &  0.090 \; (0.091)           \\				
\hline
& Category fixed effects           & {Yes} \\
& Time fixed effects               & {Yes} \\
& $Post \times$ Time fixed effects & {Yes} \\
\hline
& R$^2$                            & {0.498} \\
& Num. obs.                        & {4,598} \\
\hline
\end{tabular}
\begin{tablenotes}[para,flushleft]
{\footnotesize Note: Each event contributes one observation for the pre-period and one for the post period, resulting in a total of $2,299 \times 2 = 4,598$ observations. Time fixed effects refer to event start week and day-of-week fixed effects. We standardize all continuous covariates to ensure comparability of coefficients. $^{***}p<0.01$, $^{**}p<0.05$, $^*p<0.1$.} 
\end{tablenotes}
\end{threeparttable}
\end{table}

Based on the above discussion, exclusiveness and exogeneity conditions are likely to hold in our study context. We further provide indirect empirical evidence. If the instruments only affect focal event sales by influencing the pricing decisions of the focal event creator, then regressing focal event sales on both the instruments and the explanatory variable (i.e., focal event price) should not yield significant coefficients for the instruments. Consistent with this expectation, we find that the estimation yields insignificant coefficients on the instruments ($p > 0.05$) but significant coefficients on the explanatory variable ($p < 0.01$).
	
\subsubsection{2SLS Results}

\noindent We re-estimate Equation \eqref{eq:main} using the two-stage least squares (2SLS) approach to derive our IV estimates. \Cref{tab:iv_main} reports the results from the second-stage regression. We find that the IV estimates are similar to the OLS estimates shown in Table \ref{tab:ols_main}. This again confirms that the OLS estimates with a rich set of control variables are unlikely to suffer from omitted variable bias in our context.

For robustness checks, we follow the approach in Section~\ref{sec:robu_controls} by including category-specific time trends and allowing for fully flexible covariate effects across periods. We also estimate a specification that augments our main model with event embeddings, as noted earlier, to better account for event content. These alternative specifications with enhanced controls help mitigate potential confounding pathways between the instruments and the error term, thereby reinforcing the credibility of our instruments. The results, presented in Web Appendix \ref{app:robu_iv}, remain qualitatively unchanged after including these controls.
	
\section{Mechanisms}\label{sec:mechanism}

\noindent We now investigate the underlying mechanisms driving the observed differences in price elasticity across periods. We adopt the IV approach for all analyses in this section.

\subsection{Consumer Self-Selection}

\noindent A natural explanation is consumer self-selection: certain types of consumers prefer to make purchases in the pre-livestream period and are more price-sensitive. Given the limited purchase window before the livestream, consumers who make purchases before the event is streamed are likely to be more frequent Zhihu users, compared with those who purchase afterward. Prior research indicates that frequent consumers generally possess greater market knowledge and higher awareness of alternatives, thereby exhibiting greater price sensitivity \citep{kim1994purchase}. Moreover, frequent Zhihu users are often domain experts with higher within-platform reputations, which further reinforces their selectivity and price sensitivity when purchasing paid events.\footnote{Another possibility is that followers of a creator may have a stronger desire to engage in the creator's event and interact with her in real time. As a result, followers may be more likely to purchase their followees' events before the livestream. To investigate this, we calculate the percentage of sales generated from creator followers for each Live event, in both the pre- and post-livestream period, separately. We find that the average percentage of sales from creator followers per event is 25\% in the pre-period and 21\% in the post-period. This minor difference seems insufficient to explain the approximately 30\% difference in the scale of price elasticity across periods.}

To investigate this, we compare average consumer characteristics in the pre- and post-periods for each live event. We focus on four key engagement metrics: posting questions, posting answers, following others, and being followed. For each consumer, we calculate the cumulative activity counts at the time of purchase. As shown in \Cref{tab:cons_chars}, consumers who purchase in the pre-period exhibit significantly higher engagement levels than those who purchase in the post-period. On average, pre-period consumers have contributed 80\% more answers and 57\% more questions, and have obtained 115\% more followers and 34\% more followees compared to their post-period counterparts. These statistics confirm that pre-period purchasers are indeed more active on the platform.

	\begin{table}[!h]
		\centering
		\caption{Consumer Characteristics across Periods}
		\label{tab:cons_chars}
		\begin{threeparttable}
			\begin{tabular}{l rrrr} 
				\hline
				& Pre-livestream & Post-livestream & $t$-stats (p-value) \\
				\hline
				Num. questions & 3.44   & 2.19   & 3.96 (0.000)     \\
				Num. answers   & 24.47  & 13.61  & 13.74 (0.000)    \\
				Num. followers & 1331.72  & 620.32  & 6.69 (0.000)  \\
				Num. followees & 110.95   & 82.93   & 12.41 (0.000) \\
				% Being the creator's follower (0/1)  & 0.25  & 0.21  & 14.24 (0.000) \\
				\hline
			\end{tabular}
			\begin{tablenotes}[para,flushleft]
				{\footnotesize Note: We calculate the average consumer characteristics for each event separately in the pre- and post-livestream periods, based on 2,251 Live events as 48 events have zero sales in the post-period.}
			\end{tablenotes}
		\end{threeparttable}
	\end{table}

	\begin{table}[!h]
		\small
		\renewcommand{\arraystretch}{1}
		\addtolength{\tabcolsep}{0pt}
		\centering
		\begin{threeparttable}
			\caption{IV Estimation: Heterogeneity By Consumer Engagement Levels}\label{tab:cons_seg_reg_iv}
			\begin{tabular}{ll *{5}{S[table-format=2.3, table-space-text-post=***]}}
				\hline
				& & \multicolumn{2}{c}{\textit{Average Price Elasticity}} & &  \multicolumn{2}{c}{\textit{Elasticity Gap across Periods}} \\
				\cline{3-4} \cline{6-7} 
				&  {Dependent Variable} & {$Ln(Demand^{L}$)} & {$Ln(Demand^{H}$)} & & {$Ln(Demand^{L}$)} & {$Ln(Demand^{H}$)} \\
				& & {(1)} & {(2)} & & {(3)} & {(4)} \\
				\hline
                &{$Ln(Price)$}             & -0.884$^{***}$ & -1.018$^{***}$  &  & -1.084$^{***}$ & -1.283$^{***}$  \\
				&                          & {(0.087)}      & {(0.092)}       &  & {(0.104)}      & {(0.103)}       \\
				&{$Post$}                  &                &                 &  & -2.020$^{**}$  & -2.733$^{***}$  \\
				&                          &                &                 &  & {(0.841)}      & {(0.838)}       \\
                &{$Post \times Ln(Price)$} &                &                 &  & 0.399$^{***}$  & 0.529$^{***}$   \\
				&                          &                &                 &  & {(0.135)}      & {(0.134)}       \\
				&Event characteristics     & {Yes}          & {Yes}           &  & {Yes}          & {Yes}           \\
				&Creator characteristics   & {Yes}          & {Yes}           &  & {Yes}          & {Yes}           \\
                \hline
                &Category fixed effects    & {Yes}          & {Yes}           &  & {Yes}          & {Yes}           \\
				&Time fixed effects        & {Yes}          & {Yes}           &  & {Yes}          & {Yes}           \\
				&$Post \times$ time fixed effects  & {Yes}  & {Yes}           &  & {Yes}          & {Yes}           \\
				\hline
				&R$^2$                     &  {0.275}       &  {0.345}        &   & {0.407}       & {0.531}         \\
                &Num. obs.                 &  {4,598}       &  {4,598}        &   & {4,598}       & {4,598}         \\
				\hline
			\end{tabular}
			\begin{tablenotes}[para,flushleft]
				{\footnotesize Note: Each event contributes one observation for the pre-period and one for the post-period, resulting in a total of $2,299 \times 2 = 4,598$ observations. Time fixed effects refer to event start week and day-of-week fixed effects. We standardize all continuous covariates to ensure comparability of coefficients. Event characteristics and creator profiles in both the Q\&A community and the Zhihu Live market are omitted for ease of tabulation. $^{***}p<0.01$, $^{**}p<0.05$, $^*p<0.1$.}
			\end{tablenotes}
		\end{threeparttable}
	\end{table}

Next, we examine whether demand from consumers with higher engagement exhibits higher price sensitivity. Given the sparsity of user engagement, we construct a composite engagement metric, denoted as $Engagement$, by adding together the four user-level metrics in \Cref{tab:cons_chars}. All of the 613,650 unique consumers in our sample are divided into two equally sized groups based on the median $Engagement$ value.\footnote{2.27\% of consumers are categorized into both segments due to their time-varying engagement levels across multiple transactions.} The low-engagement group accounts for 34.83\% of transactions, while the high-engagement group accounts for the rest. For each event, we calculate the sales from the low- and high-engagement groups in each period, denoted as $Demand^{L}_{j,s}$ and $Demand^{H}_{j,s}$. We estimate the average price elasticity for each group separately, using the following specification: $Ln(Demand_{j,s}^{G}) = \alpha + \beta_{1}Ln(Price_{j}) + \gamma X_{j_{it}} + \epsilon_{j, s}, G\in\{L, H\}$. Columns (1) and (2) of \Cref{tab:cons_seg_reg_iv} report the 2SLS results. The price elasticity among the high-engagement users is slightly more negative than that among the low-engagement users ($-1.043$ versus $-0.910$). This pattern confirms that consumer self-selection (partially) explains the observed difference in price elasticity.

Nevertheless, the magnitude of this difference---approximately 13\%---appears insufficient to fully account for the nearly 30\% difference in price elasticity across periods. This suggests that additional mechanisms may be at play beyond consumer self-selection. If such mechanisms exist, we should continue to observe a temporal shift in price elasticity even after conditioning on group membership. To test this, we re-estimate Equation \eqref{eq:main} separately for each group, using 2SLS. As shown in Columns (3) and (4) of \Cref{tab:cons_seg_reg_iv}, the coefficient on $Post \times Ln(Price)$ remains positive and statistically significant within both groups ($p<0.01$). This observation suggests that consumer self-selection alone does not fully account for the observed elasticity gap.

\subsection{Quality Uncertainty}

\noindent Literature suggests that when quality is uncertain, consumers tend to exhibit greater price sensitivity to minimize potential losses \citep{ellsberg2001risk}. In our context, quality uncertainty is particularly salient in the pre-livestream period due to the experiential nature of live events and the fact that most creators host only a limited number of events. In contrast, quality uncertainty naturally declines in the post-livestream period, as the event has concluded and word-of-mouth information becomes available. One important source of such word-of-mouth is the ratings and reviews submitted by consumers upon event completion. Our text analyses in Web Appendix \ref{app:review} suggest that these reviews predominantly focus on content quality and learning experience. For example, as shown in \Cref{tab:bertopic}, the top-3 latent topics extracted from reviews using BERTopic \citep{grootendorst2022bertopic} consistently center on content quality, informativeness, and learning. These patterns suggest that such word-of-mouth serves as a useful source of quality-relevant information for prospective buyers in the post-period. This reduction in perceived quality uncertainty thus offers a plausible additional mechanism underlying the observed elasticity gap.

\begin{table}[!h]
    \footnotesize
    \renewcommand{\arraystretch}{1.2}
    \centering
    \begin{threeparttable}
        \caption{Top Keywords in Latent Topics from Event Reviews}
        \label{tab:bertopic}
        \begin{tabular}{lp{10cm}}
            \hline
            {Topic Name} & {Top 10 Keywords} \\
            \hline
            Content Quality and Practical Value & Content, Practical, Session, Information, Helpful, Instructor, Useful, Teacher, Learned, Packed \\
            Livestream Experience and Presentation & Live, Stream, Live stream, Zhihu, Presentation, Content, Session, Presenter, First, Listening \\
            Overall Evaluation and Rating Expressions & Star, Thumb, Five star, Five, Review, Sister, Bad, Give, Awesome, Master \\
            \hline
        \end{tabular}
        \begin{tablenotes}[para, flushleft]
            {\footnotesize Note: Topics are ranked by the number of reviews assigned and are named based on the semantic meaning of their associated keywords. Keywords are ordered by their relevance within each topic. The top three topics together account for over 50\% of all reviews.}
        \end{tablenotes}
    \end{threeparttable}
\end{table}

If this mechanism is at play, we expect smaller temporal gaps in price elasticity for cases where quality uncertainty is relatively low and thus the marginal benefit of post-event quality revelation is smaller. We will examine three such cases. First, seller reputation often serves as an informational cue that reduces uncertainty \citep{tellis1988price, bolton1989relationship, jin2006price, yoganarasimhan2013value}. Thus, we expect a smaller elasticity shift for events hosted by more reputable creators. Second, we distinguish between ``soft'' (e.g., Lifestyle) and ``hard'' (e.g., Science) content categories. Soft categories often involve subjective evaluations, whereas hard categories allow for more standardized quality assessments based on event descriptions and credentials \citep{bebko2000service, dai2020people}. Thus, we anticipate a smaller elasticity gap for hard categories. Third, consumers who are familiar with a creator (e.g., those who follow the creator before the time of purchase) are likely to face smaller uncertainty when evaluating event quality. As such, we expect a smaller across-period elasticity gap for transactions involving such consumers.

\subsubsection{Creator Reputation}

\noindent We approximate creator reputation using two separate metrics. The first is the ``Excellent Contributor'' badge, which is awarded to creators for consistent, high-quality answer contributions. This badge is prominently displayed alongside the user’s avatar, making it highly salient to potential consumers. Within our sample, about 18\% of events are hosted by badge recipients. The second is creator follower count, a signal of quality and popularity \citep{toubia2013intrinsic, cong2025understanding}. For each event, we obtain the creator's follower count by the event start time, denoted as $Follower_{j_{it}}$. We then divide all events into five equal-sized segments using the quintiles of $Follower_{j_{it}}$. The $1_{st}$, $2_{nd}$, $3_{rd}$, and $4_{th}$ quintiles of $F_{j_{it}}$ are 168, 1,932, 8,692, and 29,022, respectively. Segment $g$ (denoted by $F^{g}, g \in \{1, 2, 3, 4, 5\}$) is the segment that contains events within the $g$th quintile range of $Follower_{j_{it}}$. We stratify all events by creator badge achievement and follower size bins, and estimate Equation \eqref{eq:main} separately for each subgroup using 2SLS.

The results are presented in \Cref{tab:event_seg_iv}. The coefficient on the interaction term $Post \times Ln(Price)$ is positive and statistically significant for both recipient and non-recipient groups ($p < 0.01$), but the effect size is smaller in magnitude for the badge recipients (Columns (1) and (2)). This confirms that stronger reputational signals attenuate the across-period shift in price sensitivity. The follower-size segments (Columns (3) -- (7)) also exhibit a consistent pattern: events hosted by less-followed creators show larger elasticity shifts. The two lowest segments display the strongest shifts, with coefficients on $Post \times Ln(Price)$ of 0.920 for $F^{1}$ ($p < 0.1$) and 0.853 for $F^{2}$ ($p < 0.05$). In contrast, the coefficients for $F^{3}$ and $F^{4}$ are small and statistically insignificant ($0.441$ and $0.375$, $p > 0.1$), suggesting less pronounced shifts in elasticity. While the coefficient for $F^{5}$ is 0.529 ($p < 0.01$), it remains below the magnitudes observed for $F^{1}$ and $F^{2}$.

Notably, pre-period price elasticity (the coefficients on $Ln(Price)$) follows an inverted U-shape across segments. Sensitivity is highest at $F^1$ and $F^2$, lowest for the mid-tier $F^3$, and increases again for $F^4$ and $F^5$. This pattern aligns with prior literature, which suggests that low follower counts signal insufficient quality and credibility, while high follower counts can undermine perceived authenticity and approachability, leaving mid-tier creators eliciting the most favorable consumer responses \citep{wies2023finding, tian2024mega}. Crucially, perceptions of inauthenticity and unapproachability are less amenable to resolution through ex post quality revelation. This helps explain why, although price sensitivity for $F^4$ and $F^5$ in the pre-period is strong, its reduction in the post-period is substantially smaller than for $F^1$ and $F^2$.

\begin{sidewaystable}
    \small
    \renewcommand{\arraystretch}{1}
    \addtolength{\tabcolsep}{0pt}
    \centering
    \begin{threeparttable}
        \caption{IV Estimation: Heterogeneity By Creator Reputation and Content Category}\label{tab:event_seg_iv}
        \begin{tabular}{ll *{10}{S[table-format=2.3, table-space-text-post=***]}}
            \hline
            & & \multicolumn{10}{c}{Dependent Variable: $Ln(Demand)$} \\
            \cline{3-12}
            &  {Event Segments} & {Non-recipient} & {Recipient} & & {$F^1$} & {$F^2$} & {$F^3$} & {$F^4$} & {$F^5$} & {Soft} & {Hard} \\
            & & {(1)} & {(2)} & & {(3)} & {(4)} & {(5)} & {(6)} & {(7)} & {(8)} & {(9)}  \\
            \hline
            &{$Ln(Price)$}             & -1.232$^{***}$ & -1.102$^{***}$  &  & -1.395$^{***}$ & -1.495$^{***}$   & -0.556$^{**}$  & -1.484$^{***}$ & -1.502$^{***}$ & -1.295$^{***}$ & -1.147$^{***}$ \\
            &                          & {(0.132)}      & {(0.190)}       &  & {(0.358)}      & {(0.297)}        & {(0.222)}      & {(0.231)}      & {(0.175)}      & {(0.159)}      & {(0.146)} \\
            &{$Post$}                  & -2.792         & -2.537$^{***}$  &  & -3.760$^{***}$ & -2.756$^{**}$    & -2.733$^{**}$  & -2.145$^{**}$  & -3.368$^{***}$ & -2.970$^{*}$   & -2.322$^{**}$ \\
            &                          & {(1.726)}      & {(0.869)}       &  & {(1.244)}      & {(1.138)}        & {(1.156)}      & {(1.067)}      & {(1.128)}      & {(1.703)}      & {(0.971)}  \\
            &{$Post \times Ln(Price)$} & 0.579$^{***}$  & 0.402$^{***}$   &  & 0.920$^{*}$    & 0.853$^{**}$     & 0.441          & 0.375          & 0.529$^{***}$  & 0.683$^{***}$  & 0.332$^{*}$      \\
            &                          & {(0.179)}      & {(0.197)}       &  & {(0.486)}      & {(0.408)}        & {(0.283)}      & {(0.305)}      & {(0.198)}      & {(0.206)}      & {(0.183)} \\
            &Event characteristics     & {Yes}          & {Yes}           &  & {Yes}          & {Yes}            & {Yes}          & {Yes}          & {Yes}          & {Yes}          & {Yes} \\
            &Creator characteristics   & {Yes}          & {Yes}           &  & {Yes}          & {Yes}            & {Yes}          & {Yes}          & {Yes}          & {Yes}          & {Yes} \\
            \hline
            &Category fixed effects    & {Yes}          & {Yes}           &  & {Yes}          & {Yes}            & {Yes}          & {Yes}          & {Yes}          & {Yes}          & {Yes} \\
            &Time fixed effects        & {Yes}          & {Yes}           &  & {Yes}          & {Yes}            & {Yes}          & {Yes}          & {Yes}          & {Yes}          & {Yes} \\
            &$Post \times$ time fixed effects  & {Yes}  & {Yes}           &  & {Yes}          & {Yes}            & {Yes}          & {Yes}          & {Yes}          & {Yes}          & {Yes} \\
            \hline
            &R$^2$                     &  {0.500}       &  {0.658}        &   & {0.660}       & {0.586}          & {0.575}        & {0.581}        & {0.665}        & {0.502}        & {0.551} \\
            &Num. obs.                 &  {3,778}       &  {820}          &   & {920}         & {920}            & {920}          & {919}          & {919}          & {2,498}        & {2,100} \\
            \hline
        \end{tabular}
        \begin{tablenotes}[para,flushleft]
            {\footnotesize Note: $F^{g}$ ($g \in \{1, 2, 3, 4, 5\}$) is the segment of events within the $g$th quintile range of $Follower_{j_{it}}$. Within each segment, every event contributes one observation for the pre-period and one for the post-period. Time fixed effects refer to event streaming week and day-of-week fixed effects. We standardize all continuous covariates to ensure comparability of coefficients. Event characteristics and creator profiles in both the Q\&A community and the Zhihu Live market are omitted for ease of tabulation. $^{***}p<0.01$, $^{**}p<0.05$, $^*p<0.1$.}
        \end{tablenotes}
    \end{threeparttable}
\end{sidewaystable}

\subsubsection{Content Categories}

\noindent Of the 17 topics available in the Zhihu Live market, we classify nine as \textit{soft} (Internet, Psychology, Education, Travel, Lifestyle, Food \& Cuisine, Careers, Reading \& Writing, Music \& Games \& Movies), and the remaining eight, which cover relatively more objective subject matter, as \textit{hard} (Law, Business, Economics \& Finance, Healthcare, Architecture \& Design, Art History \& Appreciation, Science \& Technology, Sports). This yields 1,050 hard events and 1,249 soft events.  We expect a smaller across-period elasticity gap for hard-category events, since their subject matter permits more standardized quality assessments based on event descriptions and creator credentials \citep{dai2020people, bebko2000service}.

We re-estimate Equation \eqref{eq:main} separately for each category using 2SLS. The results are in Columns (8) and (9) of \Cref{tab:event_seg_iv}. The coefficients of the interaction term are significantly positive in both categories ($p<0.01$ and $p<0.1$), but the magnitude in the soft category is nearly double that in the hard category. This provides additional evidence for the quality uncertainty mechanism.

\subsubsection{Consumer Prior Knowledge}

\noindent Consumers may face lower quality uncertainty (even in the pre-livestream period) if they are familiar with the creator. We approximate consumer familiarity using follow status, defined by whether a consumer had followed the creator before the time of purchase. In our data, approximately 20\% of transactions are made by consumers who were followers at the time of purchase. For each Live event, we decompose period-specific demand into that generated by followers ($Demand^{F}_{j,s}$) and non-followers ($Demand^{NF}_{j,s}$), and re-estimate Equation \eqref{eq:main} for each consumer group using 2SLS.

	\begin{table}[!h]
		\small
		\renewcommand{\arraystretch}{1}
		\addtolength{\tabcolsep}{0pt}
		\centering
		\begin{threeparttable}
			\caption{IV Estimation: Heterogeneity by Consumer Follow Status}\label{tab:cons_seg_iv}
			\begin{tabular}{ll *{2}{S[table-format=2.3, table-space-text-post=***]}}
				\hline
				&  {Dependent Variable} & {$Ln(Demand^{NF}$)} & {$Ln(Demand^{F}$)} \\
				& & {(1)} & {(2)}  \\
				\hline
                &{$Ln(Price)$}             & -1.319$^{***}$ & -0.639$^{***}$  \\
				&                          & {(0.110)}      & {((0.108))}     \\
				&{$Post$}                  & -2.465$^{***}$ & -2.732$^{***}$  \\
				&                          & {(0.891)}      & {(0.876)}       \\
                &{$Post \times Ln(Price)$} & 0.488$^{***}$  & 0.354$^{**}$  \\
				&                          & {(0.142)}      & {(0.140)}     \\
				&Event characteristics     & {Yes}          & {Yes}           \\
				&Creator characteristics   & {Yes}          & {Yes}           \\
                \hline
                &Category fixed effects    & {Yes}          & {Yes}           \\
				&Time fixed effects        & {Yes}          & {Yes}           \\
				&$Post \times$ time fixed effects  & {Yes}  & {Yes}   \\
				\hline
				&R$^2$                     & {0.475}        & {0.427}        \\
                &Num. obs.                 & {4,598}        & {4,598}        \\
				\hline
			\end{tabular}
			\begin{tablenotes}[para,flushleft]
				{\footnotesize Note: Each event contributes one observation for the pre-period and one for the post-period, resulting in a total of $2,299 \times 2 = 4,598$ observations. Time fixed effects refer to event start week and day-of-week fixed effects. We standardize all continuous covariates to ensure comparability of coefficients. Event characteristics and creator profiles in both the Q\&A community and the Zhihu Live market are omitted for ease of tabulation. $^{***}p<0.01$, $^{**}p<0.05$, $^*p<0.1$.}
			\end{tablenotes}
		\end{threeparttable}
	\end{table}

\Cref{tab:cons_seg_iv} reports the results. The coefficients on the interaction term $Post \times Ln(Price)$ are significantly positive in both groups ($p<0.05$), indicating a general reduction in price sensitivity in the post-period. However, the magnitude of the coefficient is smaller for followers than for non-followers. This result supports our hypothesis that the across-period shift in elasticity is attenuated among consumers with greater prior knowledge of the creator, consistent with the quality uncertainty mechanism.

\subsubsection{Robustness Check}

\noindent To further validate the quality uncertainty mechanism, we assess whether the observed effects persist after accounting for consumer self-selection. If quality uncertainty operates independently of consumer self-selection, the results should remain robust when the consumer composition is held constant. We therefore focus on a subset of consumers who made purchases in both the pre- and post-livestream periods. This ``intersection'' subgroup accounts for 25\% of all consumers and generates 62\% of all transactions during our study period. We replicate all the previous analyses in this section using demand generated from this subgroup, thereby mitigating unobserved heterogeneity in consumer types. The results remain qualitatively unchanged (see \Cref{tab:cons_seg_iv_intersection} and \Cref{tab:event_seg_iv_intersection} in Web Appendix \ref{app:mechanism}), suggesting that quality uncertainty affects price elasticity through a channel distinct from consumer self-selection.

We acknowledge that live and recorded content differ along multiple dimensions, and quality uncertainty is unlikely to be the sole driver of the observed elasticity gap. Nonetheless, our findings provide evidence consistent with the quality uncertainty mechanism, suggesting that it is a meaningful factor shaping consumer price sensitivity in this context.

\subsection{Alternative Explanation}\label{sec:alternative_explanation}

We consider two alternative explanations that are rooted in platform display policy rather than in the intrinsic nature of the content: ranking effects and search costs. Our goal is to assess the extent to which the observed elasticity gap reflects artifacts of how events are displayed on the platform rather than genuine differences in consumer response to live versus recorded content.

The first alternative explanation is the changing event positions. As illustrated in \Cref{fig:market-layout2}, during the pre-livestream period, events are featured more upfront and gradually move closer to the top position as the livestream date approaches. After events conclude, they move to a lower-ranked section and their positions continue to drift downward over time. The changing positions may influence how consumers respond to events and which types of consumers choose to engage, potentially contributing to the observed elasticity gap. If so, the estimated gap would be contingent on the platform’s current ranking policy and subject to policy changes.

To rule out this explanation, we examine temporal variations in price elasticity within each period. If event positions were the primary driver, we should observe not only a discontinuity at the boundary of the livestream date but also a continuous, monotonic trend within the pre- and post-period, given the gradual shifts in event ranking over time. We aggregate event-level sales by week relative to the event start date and estimate separate 2SLS regressions of log-transformed weekly sales on $Ln(Price)$ for each week, controlling for all covariates specified in Equation \eqref{eq:main}. \Cref{fig:ranking_effect} in Web Appendix \ref{app:mechanism} visualizes the estimated price coefficients for the four weeks before and after the event start day. We find that, despite a sharp shift from the pre-period to the post-period, within-period elasticity remains relatively stable. This finding suggests that the elasticity gap is unlikely to be fully driven by platform ranking policy.

% A second alternative explanation is search cost. At the time of our study, no search function was available on the Zhihu Live market. Thus, it was not easy for consumers to discover both live and recorded events. Given the structure of how events were listed on the platform (see discussion above), it is possible that event ranking (position) could mitigate part of the search cost. However, other aspects of the platform interface could still be different, especially across the pre- and post-periods. For example, the non-decreasing archive size of recorded events could make it harder to discover specific events in the post-period relative to the pre-period. This could then impact the estimated price elasticities. Given that we have no ability to measure search cost directly, we acknowledge that we cannot fully rule out the search cost explanation. We hope that future research can address this.

A related alternative explanation is search cost. In the pre-period, events are prominently featured upfront, whereas in the post-period consumers must navigate an archive of recorded content, potentially leading to higher search costs. This difference may affect how consumers form consideration sets and compare alternatives across periods, thereby contributing to the observed elasticity gap. This concern is partly mitigated by our preceding discussion of position effects, since search costs are likely shaped in part by event positions (rankings), especially given that no search function was available on the Zhihu Live platform during our study period. However, because we are unable to directly measure search costs, and because other factors may also contribute to search costs (e.g., the growing archive of recorded events), we cannot fully rule out this alternative explanation. We hope future research can further examine the role of search costs in shaping the estimated price elasticity.

\section{Policy Implications}\label{sec:policy}

Moving beyond elasticity estimation, we examine how differences in price elasticities across periods affect revenue and pricing strategy. Specifically, we compare two pricing regimes -- time-invariant pricing and period-specific pricing -- by deriving the optimal prices and revenues under each. The goal of this analysis is to offer implications for platform pricing policy, given the distinct demand characteristics of live and recorded content. For ease of interpretation, prices are reported in RMB throughout this section.

\subsection{Deriving Optimal Prices}

\noindent We first consider time-invariant pricing, in which a content creator selects a single price, $P$, for an event to maximize total revenue. This requires three key inputs: 1) the features of the specific event, $X$; 2) the estimated pre- and post-period demand function $Q_{s}(P) = \alpha_{s} + \beta_{s} P + \gamma_{s}^\top X, s\in\{Pre, Post\}$\footnote{We adopt the separate period-specific regressions for simplicity and transparency. To avoid complexity due to non-linear transformations in the optimization process, we use raw-scaled price and demand instead of the log-transformed values. All of our previous findings remain consistent under this specification.}; 3) an assumption of the revenue and cost structure. During our study period, Zhihu charges 0\% commission fee. We also assume zero marginal cost per additional ticket sold, following standard assumptions in the digital goods pricing literature \citep{jain2002pricing}. We omit subscripts of all notations for simplicity.

Under these conditions, the revenue maximization problem is defined as:
\begin{align}
\max_{P} R(P) 
&= \max_{P} P \cdot (Q_{\text{Pre}} + Q_{\text{Post}}) \\
&= P \cdot \left( (\alpha_{\text{Pre}} + \beta_{\text{Pre}} P + \gamma_{\text{Pre}}^{\top} X) 
               + (\alpha_{\text{Post}} + \beta_{\text{Post}} P + \gamma_{\text{Post}}^{\top} X) \right) \\
&= P \cdot \left( 
\underbrace{\alpha_{\text{Pre}} + \alpha_{\text{Post}} + (\gamma_{\text{Pre}} + \gamma_{\text{Post}})^{\top} X}_{a_j} 
+ \underbrace{(\beta_{\text{Pre}} + \beta_{\text{Post}})}_{b} \cdot P 
\right)
\end{align}
where $a_j$ denotes the event-specific baseline demand component and $b$ is the combined price coefficient. The revenue-maximizing price $P^{*}$ is derived by solving the first-order condition: $P^* = -\frac{a_j}{2b}$. Substituting $P^*$ back into the revenue function yields the optimal revenue: $R^* = -\frac{a_j^2}{4b}$.

We next consider a hypothetical scenario with period-specific pricing, in which the creator can set different prices for each period. We make two assumptions. First, creators are sufficiently sophisticated to implement period-specific pricing strategies. Second, demand in each period is independent. The latter assumption is admittedly strong, as demand likely exhibits intertemporal dependencies. For instance, a lower pre-period price could stimulate early sales, which cannibalize post-period demand or possibly serve as a positive signal that boosts post-period demand. Because the direction and magnitude of these effects are ambiguous, we adopt an ``independent states'' assumption to maintain parsimony and avoid imposing speculative structure on the model.

For each period $s \in \{Pre, Post\}$, the optimal price and revenue are given by: $P^{*}_{s} = - \frac{a_{j, s}}{2b_{s}}$ and $R^{*}_{s} = -\frac{a_{j, s}^{2}}{4b_{s}}$, where $a_{j, s} = \alpha_{s} + \gamma_{\text{s}}^{\top} X$ denotes the baseline demand for event $j$ in period $s$ and $b_{s} = \beta_{s}$ is the estimated price coefficient from the corresponding demand function.\footnote{The derived optimal prices depend on the estimates of event-specific baseline demand (i.e., $a_j$ or $a_{j,s}$) and are therefore influenced by the accuracy of these estimates. Since our main objective is to compare optimal outcomes across pricing structures, predictive precision is not critical for this comparative analysis.}

\subsection{Time-invariant Pricing Versus Period-specific Pricing}

\begin{table}[!h]
\small
\renewcommand{\arraystretch}{1.2}
\addtolength{\tabcolsep}{0pt}
\centering
\begin{threeparttable}
\caption{Optimal Prices Under Time-invariant Pricing and Period-specific Pricing Structure}\label{tab:opt_price_rmb}
\begin{tabular}{l rrr rrr rrr}
\hline
& \multicolumn{2}{c}{$P^{*}$} &
& \multicolumn{2}{c}{$P^{*}_{Pre}$} &
& \multicolumn{2}{c}{$P^{*}_{Post}$} \\
\cline{2-3} \cline{5-6} \cline{8-9}
Segment & Mean & Std. &
        & Mean & Std. &
        & Mean & Std. \\
\hline
All Events        & 35.805 & 23.450  &  & 34.412 & 23.464 &   & 41.958 & 26.936 \\
\hline 
Non-recipient     & 35.406 & 23.345  &  & 33.768 & 23.282 &   & 42.651 & 26.922 \\
Recipient         & 37.653 & 23.877  &  & 37.394 & 24.087 &   & 38.794 & 26.803 \\
\hline 
$F_1$            & 26.320 & 18.942  &  & 24.927 & 19.390 &   & 32.501 & 21.070 \\
$F_2$            & 30.219 & 19.019  &  & 28.518 & 19.222 &   & 37.716 & 21.924 \\
$F_3$            & 35.476 & 21.084  &  & 34.055 & 21.098 &   & 41.755 & 24.472 \\
$F_4$            & 41.363 & 24.472  &  & 40.005 & 24.311 &   & 47.355 & 28.273 \\
$F_5$            & 45.675 & 27.202  &  & 44.583 & 26.754 &   & 50.505 & 33.187 \\
\hline 
Soft Category    & 33.810 & 23.093  &  & 32.599 & 23.156 &   & 39.165 & 25.921 \\
Hard Category    & 38.185 & 23.660  &  & 36.575 & 23.653 &   & 45.290 & 27.741 \\
\hline
\end{tabular}
\begin{tablenotes}[para,flushleft]
\footnotesize
Note: The number of observations is 2,299 Live events. $F^{g}$ ($g \in \{1, 2, 3, 4, 5\}$) is the segment of events within the $g$th quintile range of $Follower_{j_{it}}$. The unit of price is RMB.
\end{tablenotes}
\end{threeparttable}
\end{table}

\noindent For each event, we compute the optimal prices under each pricing policy. Table \ref{tab:opt_price_rmb} reports the average optimal prices across all events and by event segments defined by follower-size quintiles, badge status, and topical category. On average, the optimal time-invariant price lies between the two period-specific prices, i.e., $P^{*}$ is higher than $P^{*}_{Pre}$ but lower than $P^{*}_{Post}$. The average difference between $P^{*}_{Pre}$ and $P^{*}_{Post}$ is approximately $(41.958 - 34.412)/34.412 = 21.93\%$. This pattern holds consistently across all event segments.

\begin{figure}[h!]
    \centering
    \includegraphics[height=3.5in]{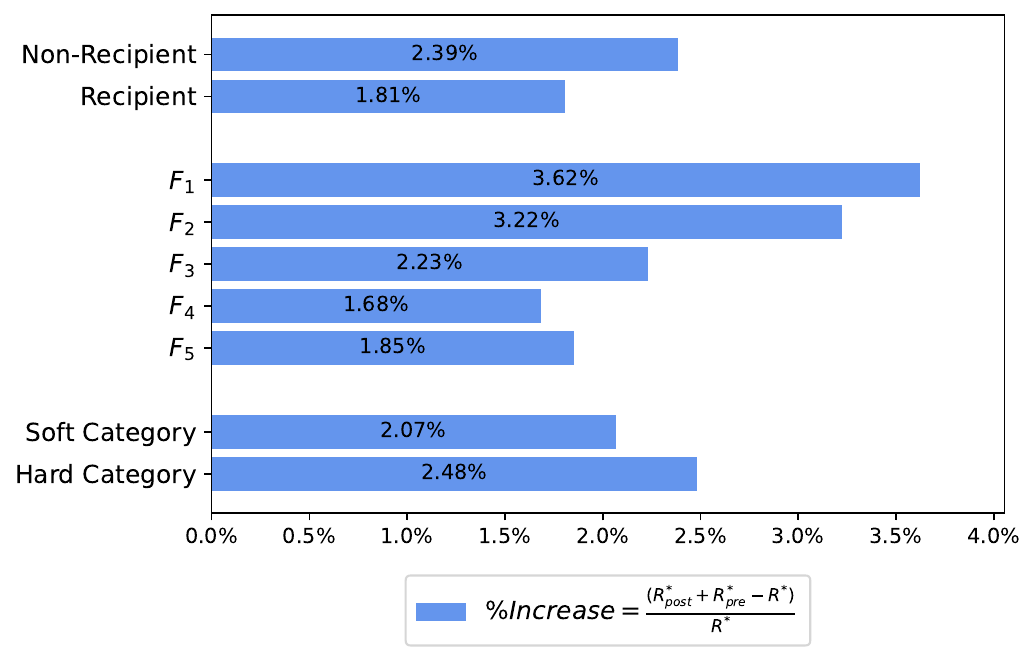}
    \caption{Percentage Revenue Gain from Period-specific Pricing}
    \label{fig:stats_counterfactual_dynamic}
    \begin{minipage}{0.9\textwidth}
        {\footnotesize Note: This plot presents the average percentage increase in optimal revenue of period-specific pricing relative to time-invariant pricing across events within each segment. $F^{g}$ ($g \in \{1, 2, 3, 4, 5\}$) is the segment of events within the $g$th quintile range of $Follower_{j_{it}}$.}
    \end{minipage}
\end{figure}

We further derive the corresponding optimal revenues. For each event, we calculate the percentage increase in optimal revenue from adopting period-specific pricing relative to time-invariant pricing: $ \%Increase = \frac{(R^{*}_{Post} + R^{*}_{Pre} - R^{*})}{R^{*}}$. On average, period-specific pricing yields a 2.28\% increase in total revenue compared to time-invariant pricing. \Cref{fig:stats_counterfactual_dynamic} shows these revenue gains by segment. Notably, creators with smaller follower bases benefit more from period-specific pricing: creators in $F_1$ experience an average revenue increase of approximately 3.6\%, compared to only 1.8\% for creators in $F_5$. Similarly, creators without the ``Excellent Contributor'' badge tend to gain more than badge recipients. Viewed through the lens of the quality uncertainty mechanism, period-specific pricing confers greater advantages for creators whose audiences face higher ex ante uncertainty about event quality.

An important caveat is that these policy simulations rest on assumptions of creator sophistication and demand independence across periods. We acknowledge that alternative assumptions may yield different comparative outcomes between time-invariant and period-specific pricing. Despite this limitation, the exercise links differences in period-specific price elasticities to revenue implications and pricing strategies, and offers insight into the potential benefits of implementing period-specific pricing.

\section{Conclusion}\label{sec:conclusion}
	
\noindent This paper empirically examines consumer price sensitivity for paid livestreaming events before and after the livestream, using data from a large knowledge-sharing platform. This platform allows creators to launch pay-per-view livestreaming events and makes the recorded versions available at the same price after events are over. We find that the price elasticity of demand is more price-sensitive in the pre-livestream period, indicating a surprisingly higher willingness to pay for recorded versions. Further analyses show that the observed shifts in price elasticity across periods can be partly explained by two mechanisms: consumer self-selection and quality uncertainty.
	
Our paper contributes to the literature and practice of the livestreaming economy by examining the complete lifecycle of livestreaming events, from the pre-livestream period to the subsequent recorded format. This provides a holistic view of consumer engagement in the livestreaming markets and reveals the inherent nature of this type of media format. Our findings suggest that although many platforms and creators currently focus their promotional efforts on the pre-livestream period, there is substantial revenue potential in the post-livestream period. Platforms may consider rebalancing their marketing strategies to better leverage the higher willingness to pay observed during the post-period window. In addition, our policy simulations suggest the potential benefits of implementing period-specific pricing in this setting, under assumptions about creator sophistication and intertemporal demand structure. We believe that our findings have broader implications, given the widespread adoption of livestreaming media globally.

This research has several limitations, which open potential directions for future research. First, our analysis is based on data from a single platform where creators offer paid live events focused primarily on knowledge-sharing rather than entertainment. Future research could extend this investigation to other platforms such as Twitch, where providing entertainment is a major objective for creators. Second, we do not have data regarding how consumers engage with the content (e.g., minutes watched, pausing, and repeat watching of recorded content). Access to such data can help understand the drivers of consumer consumption in more detail. Third, the creators in our sample are compensated via direct pricing. Livestreamers who are compensated through indirect pricing (e.g., via brand sponsorships and paid subscriptions) may deliver content and engage with consumers during live events in different ways.

\section*{Funding and Competing Interests}
	
\noindent All authors certify that they have no affiliations with or involvement in any organization with any financial or non-financial interest in the subject matter or materials discussed in this manuscript. The research is not supported by any funding institution. 

% \noindent \textbf{Funding and Competing Interests:} All authors certify that they have no affiliations with or involvement in any organization or entity with any financial interest or non-financial interest in the subject matter or materials discussed in this manuscript. 

% \subsection*{Acknowledgements}

%  \noindent M. Yin acknowledges the support from the Marketing Science Institute and the Warrington Commitment Research Award. J. Liu acknowledges the support of the University Grants Committee of the Hong Kong Special Administrative Region, Hong Kong, China (Project No. 16500124, 2024-27). The authors acknowledge the computational support from UF Research Computing and HiPerGator.  

%%%%-----------------------------reference--------------------------------%%%%
%\setstretch{1.0}
%\restoregeometry
%\newgeometry{margin=1in}
%\singlespacing
\bibliographystyle{jmr.bst}
\bibliography{pricing}

%%%%-----------------------------appendix--------------------------------%%%%
% \clearpage
% \appendix
	
% \setcounter{figure}{0}
% \renewcommand{\thefigure}{A\arabic{figure}}
% \setcounter{table}{0}
% \renewcommand{\thetable}{A\arabic{table}}
% \setcounter{footnote}{0}
	
% \begin{center}
% {\Large\textit{APPENDIX}}
% \end{center}
	
% \section{Zhihu Interface}\label{app:zhihu}
	
% \begin{figure}[!h]
% \centering
% \includegraphics[height=6in]{zhihu-user-profile.png}
% \caption{Example of User Profile Page}\label{fig:profile-page}
% \end{figure}
	
% \begin{figure}[!h]
% \centering
% \includegraphics[height=6in]{zhihu-live.png}
% \caption{Example of a Live Event}\label{fig:live-session}
% \begin{minipage}{0.9\textwidth} % choose width suitably
% {\footnotesize Note: This figure illustrates the interface of a Live event. During the livestream, the creator will give a real-time talk via voice, text, and picture messages. The creator can also respond to questions raised by the audience.}
% \end{minipage}
% \end{figure}

%%%%-----------------------------web appendix-----------------------------------------%%%%
\clearpage
\appendix
	
\setcounter{figure}{0}
\renewcommand{\thefigure}{A\arabic{figure}}
\setcounter{table}{0}
\renewcommand{\thetable}{A\arabic{table}}
\setcounter{footnote}{0}
	
\begin{center}
{\Large\textit{WEB APPENDIX}}
\end{center}

\section{Zhihu Interface}\label{app:zhihu}
	
\begin{figure}[!h]
\centering
\includegraphics[height=6in]{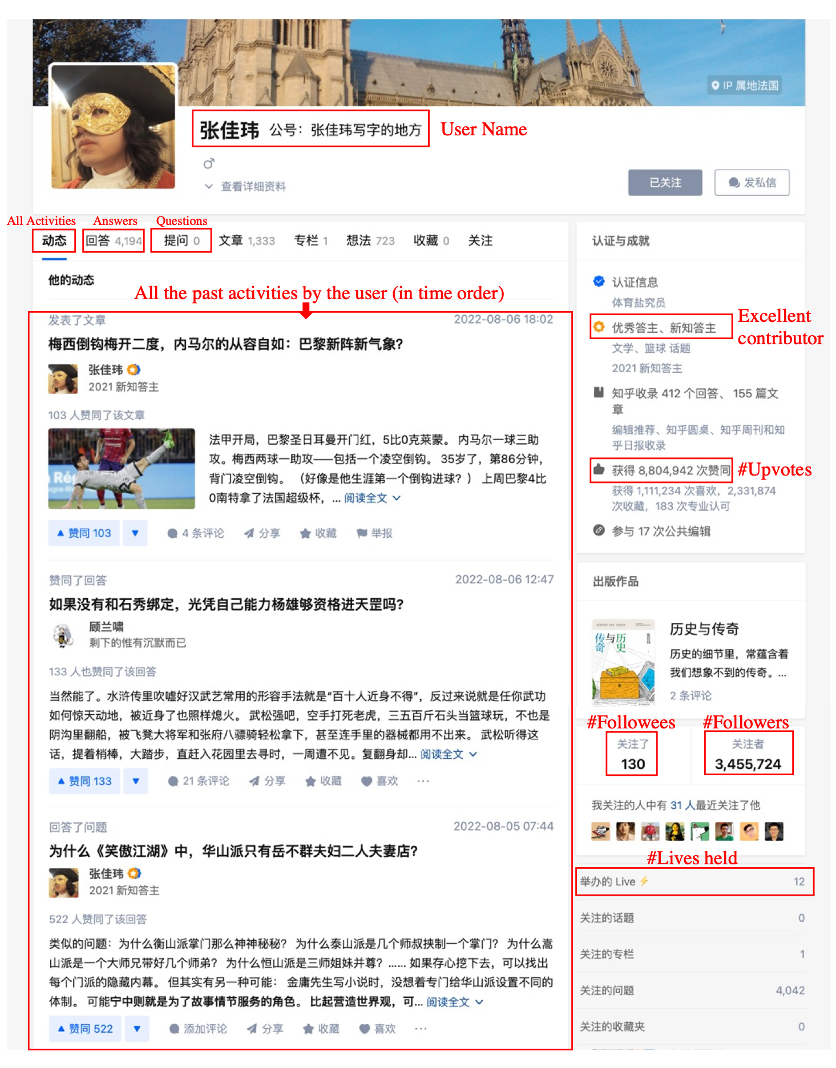}
\caption{Example of User Profile Page}\label{fig:profile-page}
\end{figure}
	
\begin{figure}[!h]
\centering
\includegraphics[height=6in]{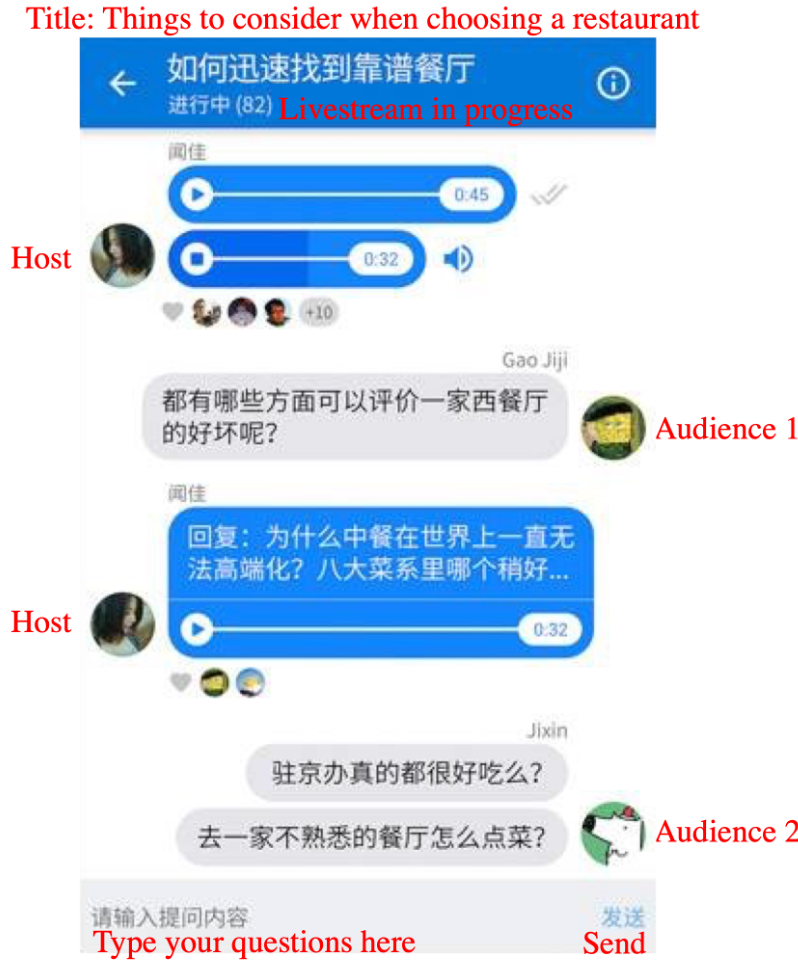}
\caption{Example of a Live Event}\label{fig:live-session}
\begin{minipage}{0.9\textwidth} % choose width suitably
{\footnotesize Note: This figure illustrates the interface of a Live event. During the livestream, the creator will give a real-time talk via voice, text, and picture messages. The creator can also respond to questions raised by the audience.}
\end{minipage}
\end{figure}

\clearpage
\newpage
\section{Price Comparison Across Comparable Content Platforms}\label{app:comp-price}

\noindent To the best of our knowledge, Zhihu Live was the largest livestreaming platform of its kind during our study period, while competing platforms remained relatively small. We identified four comparable platforms in China that may compete with Zhihu Live, including Himalaya FM, Dedao, Lychee Live, and Qianliao Live. In the summer of 2016, we conducted a survey of event prices on these five platforms by randomly scraping 100 paid events from each platform. The descriptive statistics for these prices are presented in Table \ref{tab:comp-price}. To ensure comparability across platforms, we standardize the price to USD per 30 minutes of content. The results show that paid events hosted on Zhihu generally had higher price points compared to similar events on competing platforms.

\begin{table}[!h] 
	\small 
	\renewcommand{\arraystretch}{1.2} 
	\addtolength{\tabcolsep}{0pt} 
	\centering 
	\begin{threeparttable} 
		\caption{Price Comparison (Unit: USD per 30 minutes of content)}\label{tab:comp-price} 
		\begin{tabular}{lccc} 
			\hline {Platform} & {Mean} & {Median} & {Std.} \\ 
			\hline \textbf{Zhihu Live} & \textbf{1.83} & \textbf{1.97} & \textbf{1.36} \\ 
			Himalaya FM    & 0.85 & 0.58 & 0.79 \\ 
			Dedao          & 0.50 & 0.38 & 0.37 \\ 
			Lychee Live    & 1.84 & 0.71 & 4.20 \\ 
			Qianliao Live  & 0.48 & 0.42 & 0.43 \\ 
			\hline 
		\end{tabular} 
	\end{threeparttable} 
\end{table}

\clearpage
\newpage
\section{Robustness Checks: OLS Estimation}\label{app:robu_ols}

\begin{table}[!h]
\footnotesize
\renewcommand{\arraystretch}{1}
\addtolength{\tabcolsep}{0pt}
\centering
\begin{threeparttable}
\caption{OLS Estimation: Inclusion of Additional Controls}\label{tab:ols_add_control}
\begin{tabular}{ll *{3}{S[table-format=2.3, table-space-text-post=*********]}}
\hline
\multicolumn{2}{l}{Dependent Variable: $Ln(Demand)$} & {\textit{Category-specific Time Trends}} & & {\textit{Period-specific Covariate Effects}}  \\
 & & {(1)} & & {(2)} \\
\hline
\multicolumn{5}{l}{\textit{Key Independent Variables}} \\
&{\textbf{$Post$}}                    & -2.384$^{***}$ \; (0.142)  & & -2.386$^{***}$ \; (0.891)   \\
&{\textbf{$Ln(Price)$}}               & -1.163$^{***}$ \; (0.065)  & & -1.177$^{***}$ \; (0.060)   \\
&{\textbf{$Post \times Ln(Price)$}}   &  0.329$^{***}$ \; (0.041)  & &  0.381$^{***}$ \; (0.085)   \\
\multicolumn{5}{l}{\textit{Event Characteristics}} \\
&{ExposureDuration}                   &  0.005 \; (0.048)          & & 0.029 \; (0.035)             \\
&{SeatLimit}                          & -0.094$^{*}$ \; (0.055)    & & -0.070$^{***}$ \; (0.027)    \\
&{Advertisement}                      & 0.532$^{***}$ \; (0.101)   & & 0.686$^{***}$ \; (0.111)     \\
&{Rating}                             & 0.482$^{***}$ \; (0.033)   & & 0.443$^{***}$ \; (0.037)     \\
&{Holiday}                            & -0.505$^{***}$ \; (0.143)  & & -0.358$^{***}$ \; (0.129)    \\
\multicolumn{5}{l}{\textit{Creator Profile in QA Community}} \\			
&{Follower}                           & 0.127$^{***}$ \; (0.047)   & & 0.117$^{***}$ \; (0.042)    \\
&{Followee}                           & -0.001 \; (0.023)          & & 0.039 \; (0.028)            \\
&{Answer}                             & 0.033 \; (0.029)           & & 0.037 \; (0.032)            \\
&{Upvote}                             & 0.376$^{***}$ \; (0.082)   & & 0.356$^{***}$ \; (0.099)    \\
&{Downvote}                           & 0.205$^{***}$ \; (0.064)   & & 0.184$^{***}$ \; (0.062)    \\
&{Thank}                              & 0.005 \; (0.121)           & & -0.046 \; (0.108)           \\
&{Unhelpful}                          & -0.412$^{***}$ \; (0.129)  & & -0.318$^{**}$ \; (0.125)   \\
&{Badge}                              & -0.042 \; (0.083)          & & -0.117 \; (0.081)           \\
&{TenureQA}                           & 0.101$^{***}$ \; (0.033)   & & 0.090$^{***}$ \; (0.030)    \\
&{Celebrity}                          & 0.850$^{***}$ \; (0.208)   & & 1.182$^{***}$ \; (0.190)    \\
\multicolumn{5}{l}{\textit{Creator Profile in Zhihu Live Market}} \\
&{NumPastEvent}                       & -0.147$^{***}$ \; (0.051)  & & -0.193$^{***}$ \; (0.043)   \\
&{AvgPastSales}                       & 0.309$^{***}$ \; (0.063)   & & 0.255$^{***}$ \; (0.055)    \\
&{AvgPastRating}                      & 0.066$^{*}$ \; (0.037)     & & 0.057$^{*}$ \; (0.033)     \\
&{TenureMarket}                       & -0.013 \; (0.053)          & & 0.032 \; (0.050)            \\
&{Invitation}                         & 0.076 \; (0.142)           & & 0.122 \; (0.127)            \\
\hline
& Category fixed effects                                        & {Yes}  & & {Yes} \\
& Time fixed effects                                            & {Yes}  & & {Yes} \\
& Category $\times$ Time fixed effects                          & {Yes}  & & {No}  \\
& $Post$ $\times$ Category fixed effects                        & {No}   & & {Yes}  \\
& $Post$ $\times$ Time fixed effects                            & {Yes}  & & {Yes} \\
& $Post$ $\times$ Event and creator characteristics             & {No}   & & {Yes}  \\
\hline 
&R$^2$                                                          & {0.585}  & & {0.502}  \\
&Num. obs.                                                      & {4,598}  & & {4,598}  \\
\hline
\end{tabular}
\begin{tablenotes}[para,flushleft]
{\footnotesize Note: Each event contributes one observation for the pre-period and one for the post-period, resulting in a total of $2,299 \times 2 = 4,598$ observations. Time fixed effects refer to event start week and day-of-week fixed effects. We standardize all continuous covariates to ensure comparability of coefficients. $^{***}p<0.01$, $^{**}p<0.05$, $^*p<0.1$.}
\end{tablenotes}
\end{threeparttable}
\end{table}

\begin{sidewaystable}
	\footnotesize
	\renewcommand{\arraystretch}{1}
	\addtolength{\tabcolsep}{0pt}
	\centering
	\begin{threeparttable}
		\caption{OLS Estimation: Separate Pre- and Post-Period Regressions}\label{tab:ols_separate}
		\begin{tabular}{ll *{7}{S[table-format=2.3, table-space-text-post=*********]}}
		\hline
			& & \multicolumn{2}{c}{\textit{No Controls}} & & \multicolumn{2}{c}{\textit{Full Controls}} & & \multicolumn{1}{c}{\textit{Exclusion of $Rating$}}\\
			\cline{3-4}  \cline{6-7} \cline{9-9}
			& {Dependent Variable}   & {$Ln(Demand_{Pre})$}  & {$Ln(Demand_{Post})$} & &  {$Ln(Demand_{Pre})$} & {$Ln(Demand_{Post})$} & & {$Ln(Demand_{Pre})$} \\
			&  & {(1)}   & {(2)} & &  {(3)}  & {(4)}  & & {(5)} \\
			\hline
            \multicolumn{7}{l}{\textit{Key Independent Variables}} \\
            &{\textbf{$Ln(Price)$}}                           &  -0.989$^{***}$ \; (0.066) & -0.707$^{***}$ \; (0.068)  & & -1.177$^{***}$ \; (0.057)   & -0.795$^{***}$ \; (0.063)   & & -1.194$^{***}$ \; (0.059) \\
			\multicolumn{7}{l}{\textit{Event Characteristics}} \\
			&{ExposureDuration}                               &                            &                            & & 0.029 \; (0.033)            & -0.037 \; (0.033)           & & 0.063$^{*}$ \; (0.034) \\
			&{SeatLimit}                                      &                            &                            & & -0.070$^{***}$ \; (0.026)   & -0.057$^{**}$ \; (0.028)    & & -0.070$^{***}$ \; (0.026) \\
			&{Advertisement}                                  &                            &                            & & 0.686$^{***}$ \; (0.105)    &  0.422$^{***}$ \; (0.117)   & & 0.714$^{***}$ \; (0.109) \\
			&{Rating}                                         &                            &                            & & 0.443$^{***}$ \; (0.035)    &  0.495$^{***}$ \; (0.038)   & & \\
			&{Holiday}                                        &                            &                            & & -0.358$^{***}$ \; (0.122)   & -0.425$^{***}$ \; (0.135)   & & -0.340$^{***}$ \; (0.126) \\
			\multicolumn{7}{l}{\textit{Creator Profile in QA Community}} \\			
			&{Follower}                                       &                            &                            & & 0.117$^{***}$ \; (0.040)    & 0.097$^{**}$ \; (0.044)    & & 0.110$^{**}$ \; (0.041) \\
			&{Followee}                                       &                            &                            & & 0.039 \; (0.026)            &  0.010 \; (0.029)          & & 0.041 \; (0.027) \\
			&{Answer}                                         &                            &                            & & 0.037 \; (0.030)            &  0.016 \; (0.033)          & & 0.037 \; (0.031) \\
			&{Upvote}                                         &                            &                            & & 0.356$^{***}$ \; (0.094)    &  0.439$^{***}$ \; (0.104)  & & 0.399$^{***}$ \; (0.097) \\
			&{Downvote}                                       &                            &                            & & 0.184$^{***}$ \; (0.059)    &  0.157$^{**}$ \; (0.065)   & & 0.172$^{***}$ \; (0.061) \\
			&{Thank}                                          &                            &                            & & -0.046 \; (0.102)           &  -0.061 \; (0.113)         & & -0.098 \; (0.105) \\
			&{Unhelpful}                                      &                            &                            & & -0.318$^{***}$ \; (0.119)   &  -0.362$^{***}$ \; (0.132) & & -0.291$^{**}$ \; (0.123) \\
			&{Badge}                                          &                            &                            & & -0.117 \; (0.077)           &  -0.145$^{*}$ \; (0.085)   & & -0.042 \; (0.079) \\
			&{TenureQA}                                       &                            &                            & & 0.090$^{***}$ \; (0.028)    &  0.079$^{**}$ \; (0.031)   & & 0.096$^{***}$ \; (0.029) \\
			&{Celebrity}                                      &                            &                            & & 1.182$^{***}$ \; (0.180)    &  0.354$^{*}$ \; (0.199)    & & 1.112$^{***}$ \; (0.186) \\
			\multicolumn{7}{l}{\textit{Creator Profile in Zhihu Live Market}} \\
			&{NumPastEvent}                                   &                            &                            & & -0.193$^{***}$ \; (0.041)   &  -0.084$^{*}$ \; (0.045)   & & -0.227$^{***}$ \; (0.042) \\
			&{AvgPastSales}                                   &                            &                            & & 0.255$^{***}$ \; (0.052)    &  0.354$^{***}$ \; (0.057)  & & 0.264$^{***}$ \; (0.054) \\
			&{AvgPastRating}                                  &                            &                            & & 0.057$^{*}$ \; (0.031)      &  0.009 \; (0.035)          & & 0.099$^{***}$ \; (0.032) \\
			&{TenureMarket}                                   &                            &                            & & 0.032 \; (0.047)            &  -0.034 \; (0.053)         & & 0.027 \; (0.049) \\
			&{Invitation}                                     &                            &                            & & 0.122 \; (0.120)            &  0.044 \; (0.133)          & & 0.198 \; (0.124) \\
			\hline
            &Category fixed effects                           & {Yes}                      & {Yes}                      & & {Yes}                       & {Yes}                      & & {Yes} \\
			&Time fixed effects                               & {Yes}                      & {Yes}                      & & {Yes}                       & {Yes}                      & & {Yes} \\
			\hline 
			&R$^2$                                            & {0.090}                    & {0.045}                    & & {0.461}                     & {0.352}                    & & {0.421}  \\
            &Num. obs.                                        & {2,299}                    & {2,299}                    & & {2,299}                     & {2,299}                    & & {2,299}  \\
			\hline
		\end{tabular}
		\begin{tablenotes}[para,flushleft]
		{\footnotesize Note: Time fixed effects refer to event start week and day-of-week fixed effects. We standardize all continuous covariates to ensure comparability of coefficients. $^{***}p<0.01$, $^{**}p<0.05$, $^*p<0.1$.}
		\end{tablenotes}
	\end{threeparttable}
\end{sidewaystable}

\begin{figure}[!h]
\centering
\includegraphics[height=3in]{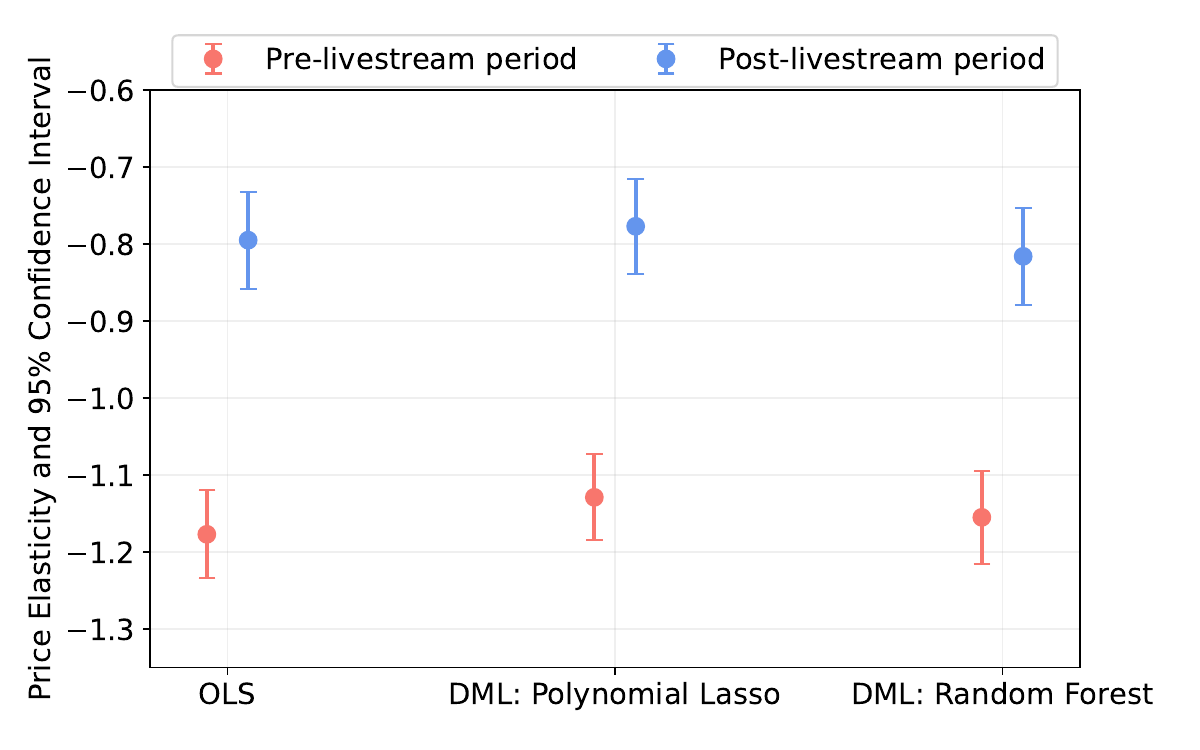}
\caption{Comparing Estimated Price Elasticities across Functional Forms}\label{fig:dml}
    \vspace{2mm} % optional spacing
    \begin{minipage}{0.7\textwidth}
        \footnotesize Note: We compare the OLS estimates (Columns (3) and (4) of \Cref{tab:ols_separate}) with the DML estimates for the pre- and post-periods. For DML, we consider two candidate models for nuisance estimation: Lasso and Random Forest.
    \end{minipage}
\end{figure}

\clearpage
\newpage
\section{Double Machine Learning}\label{app:DML}

\noindent Double Machine Learning (DML) \citep{chernozhukov2018double} is a method for estimating treatment effects when potential confounders/controls are either too many (high-dimensional) or their effects on the treatment and outcome cannot be satisfactorily modeled by parametric functions.

Consider the following partially linear regression (PLR) models:
\begin{align}
	& Y = \theta T + g(X) + \epsilon, \quad  \mathbb{E}[\epsilon|X] = 0 \\
	& T = f(X) + \eta, \quad  \mathbb{E}[\eta|X] = 0
\end{align}
where $Y$ is the outcome variable (e.g., demand), $T$ is the treatment variable (e.g., price), and $X$ is the (high-dimensional) confounding covariates. Our goal is to estimate the average treatment effect $\theta$.

We can rewrite the partially linear equations as
\begin{align}
	Y - \mathbb{E}[Y|X] = \theta(T - \mathbb{E}[T|X]) + \epsilon.
\end{align}	
Thus, if we can estimate the conditional expectation functions (both of which are pure predictive tasks): $q(X) = \mathbb{E}[Y|X]$\footnote{Note that we do not estimate $g(.)$ directly, but the conditional expectation of Y given X, denoted as $q(X) = \mathbb{E}[Y|X]$.} and $f(X) = \mathbb{E}[T|X]$, then we can compute the residuals: $\tilde{Y} = Y - q(X)$ and $\tilde{T} = T - f(X) = \eta$, which are subsequently related by the equation: $\tilde{Y} = \theta \tilde{T} + \epsilon$. Consequently, estimating $\theta$ is a final regression problem, i.e.,
\begin{align}
	& \hat{\theta} = argmin_{\theta \in \Theta} \mathbb{E}[ (\tilde{Y} - \theta \tilde{T})^{2}].
\end{align}

The main advantage of DML is that it can achieve fast estimation rates and asymptotic normality on the second stage estimate $\hat{\theta}$, even if the first stage estimates on $q(X)$ and $f(X)$ are only $N^{1/4}$ consistent, due to the robustness provided by Neyman orthogonality and cross-fitting. This robustness is beneficial because it allows the final estimator for $\theta$ to perform well, even if the nuisance parameters $g(.)$ and $f(.)$ are not correctly specified or suffer from bias induced by regularization and over-fitting \citep{chernozhukov2018double,ellickson2022estimating}.

We implement DML to estimate price elasticity separately for the pre- and post-periods, considering two candidate models for nuisance estimation: Lasso and Random Forest. For Lasso, we include quadratic terms for continuous covariates to capture potential non-linearity. We use grid search with cross-validation to select optimal hyper-parameters for the first-stage models, evaluated based on Mean Squared Error (MSE). For Lasso, we focus on selecting the optimal regularization parameter, $\alpha$, which controls the strength of the penalty applied to the model’s coefficients. For Random Forest, we tune key parameters, including n\textunderscore estimators  (the number of trees), max\textunderscore depth (the maximum depth of each tree), min\textunderscore samples\textunderscore leaf (the minimum number of samples required at each leaf node), and max\textunderscore features (the maximum number of features considered at each split). All 2,299 observations are used for this tuning process. The selected hyper-parameters are then passed to the DML framework to perform the first-stage estimation.

\clearpage
\newpage
\section{Robustness Checks: IV Estimation}\label{app:robu_iv}

\begin{table}[!h]
\footnotesize
\renewcommand{\arraystretch}{1}
\addtolength{\tabcolsep}{0pt}
\centering
\begin{threeparttable}
\caption{IV Estimation: Alternative Embedding Technique and Instrument Construction Method}
\begin{tabular}{ll *{3}{S[table-format=2.3, table-space-text-post=*********]}}
\hline
\multicolumn{2}{l}{Dependent Variable: $Ln(Demand)$} & {\textit{Alternative Embedding Method}} & & {\textit{Alternative IV: Avg. Similar Event Price}}  \\
 & & {(1)} & & {(2)} \\
\hline
\multicolumn{5}{l}{\textit{Key Independent Variables}} \\
&{\textbf{$Post$}}                    & -2.560$^{***}$ \; (0.865)  & & -2.568$^{***}$ \; (0.860)   \\
&{\textbf{$Ln(Price)$}}               & -1.277$^{***}$ \; (0.122)  & & -1.255$^{***}$ \; (0.099)   \\
&{\textbf{$Post \times Ln(Price)$}}   &  0.495$^{***}$ \; (0.155)  & &  0.503$^{***}$ \; (0.130)   \\
\multicolumn{5}{l}{\textit{Event Characteristics}} \\
&{ExposureDuration}                   & -0.001 \; (0.025)          & & -0.003 \; (0.025)            \\
&{SeatLimit}                          & -0.063$^{***}$ \; (0.019)  & & -0.063$^{***}$ \; (0.019)    \\
&{Advertisement}                      & 0.547$^{***}$ \; (0.080)   & & 0.551$^{***}$ \; (0.079)     \\
&{Rating}                             & 0.469$^{***}$ \; (0.026)   & & 0.469$^{***}$ \; (0.026)     \\
&{Holiday}                            & -0.395$^{***}$ \; (0.091)  & & -0.393$^{***}$ \; (0.091)    \\
\multicolumn{5}{l}{\textit{Creator Profile in QA Community}} \\			
&{Follower}                           & 0.110$^{***}$ \; (0.030)   & & 0.108$^{***}$ \; (0.030)    \\
&{Followee}                           & 0.024 \; (0.020)           & & 0.024 \; (0.020)            \\
&{Answer}                             & 0.026 \; (0.022)           & & 0.027 \; (0.022)            \\
&{Upvote}                             & 0.396$^{***}$ \; (0.070)   & & 0.397$^{***}$ \; (0.070)    \\
&{Downvote}                           & 0.172$^{***}$ \; (0.044)   & & 0.171$^{***}$ \; (0.044)    \\
&{Thank}                              & -0.052 \; (0.076)          & & -0.053 \; (0.076)           \\
&{Unhelpful}                          & -0.341$^{***}$ \; (0.089)  & & -0.340$^{**}$ \; (0.089)    \\
&{Badge}                              & -0.130$^{**}$ \; (0.057)   & & -0.131$^{**}$ \; (0.057)    \\
&{TenureQA}                           & 0.084$^{***}$ \; (0.021)   & & 0.084$^{***}$ \; (0.021)    \\
&{Celebrity}                          & 0.781$^{***}$ \; (0.137)   & & 0.773$^{***}$ \; (0.136)    \\
\multicolumn{5}{l}{\textit{Creator Profile in Zhihu Live Market}} \\
&{NumPastEvent}                       & -0.134$^{***}$ \; (0.032)  & & -0.137$^{***}$ \; (0.031)   \\
&{AvgPastSales}                       & 0.303$^{***}$ \; (0.039)   & & 0.304$^{***}$ \; (0.039)    \\
&{AvgPastRating}                      & 0.035 \; (0.024)           & & 0.034 \; (0.024)            \\
&{TenureMarket}                       & -0.004 \; (0.036)          & & -0.002 \; (0.036)           \\
&{Invitation}                         & 0.092 \; (0.091)           & & 0.087 \; (0.090)            \\
\hline
& Category fixed effects                                        & {Yes}  & & {Yes} \\
& Time fixed effects                                            & {Yes}  & & {Yes} \\
& $Post$ $\times$ Time fixed effects                            & {Yes}  & & {Yes} \\
\hline 
&R$^2$                                                          & {0.498}  & & {0.498}  \\
&Num. obs.                                                      & {4,598}  & & {4,598}  \\
\hline
\end{tabular}
\begin{tablenotes}[para,flushleft]
{\footnotesize Note: Column (1) uses OpenAI's pre-trained embedding models to construct the IV. Column (2) uses the average price of the three most similar events as the IV. Each event contributes one observation for the pre-period and one for the post-period, resulting in a total of $2,299 \times 2 = 4,598$ observations. Time fixed effects refer to event start week and day-of-week fixed effects. We standardize all continuous covariates to ensure comparability of coefficients. $^{***}p<0.01$, $^{**}p<0.05$, $^*p<0.1$.}
\end{tablenotes}
\end{threeparttable}
\end{table}

\begin{table}[!h]
\footnotesize
\renewcommand{\arraystretch}{1}
\addtolength{\tabcolsep}{0pt}
\centering
\begin{threeparttable}
\caption{IV Estimation: Inclusion of Additional Controls}\label{tab:IV_additional_controls}
\begin{tabular}{ll *{3}{S[table-format=2.3, table-space-text-post=*********]}}
\hline
\multicolumn{2}{l}{Dependent Variable: $Ln(Demand)$} & {\textit{Category-specific}} & {\textit{Period-specific}} & {\textit{Event Embeddings}}  \\
\multicolumn{2}{l}{} & {\textit{Time Trends}} & {\textit{Covariate Effects}} & {\textit{}}  \\
 & & {(1)} & {(2)} & {(3)} \\
\hline
\multicolumn{5}{l}{\textit{Key Independent Variables}} \\
&{\textbf{$Post$}}                    & -2.565$^{***}$ \; (0.826)  & -2.590$^{***}$ \; (0.903)  & -2.565$^{***}$ \; (0.817)  \\
&{\textbf{$Ln(Price)$}}               & -1.205$^{***}$ \; (0.109)  & -1.311$^{***}$ \; (0.114)  & -1.420$^{***}$ \; (0.106)  \\
&{\textbf{$Post \times Ln(Price)$}}   &  0.500$^{***}$ \; (0.132)  &  0.582$^{***}$ \; (0.161)  &  0.500$^{***}$ \; (0.131)  \\
\multicolumn{5}{l}{\textit{Event Characteristics}} \\
&{ExposureDuration}                   &  0.002 \; (0.027)          & 0.037 \; (0.035)           & -0.009 \; (0.024)          \\
&{SeatLimit}                          & -0.094$^{***}$ \; (0.021)  & -0.070$^{***}$ \; (0.027)  & -0.051$^{***}$ \; (0.018)  \\
&{Advertisement}                      & 0.539$^{***}$ \; (0.085)   & 0.663$^{***}$ \; (0.113)   &  0.711$^{***}$ \; (0.077)  \\
&{Rating}                             & 0.482$^{***}$ \; (0.028)   & 0.442$^{***}$ \; (0.037)   &  0.454$^{***}$ \; (0.025)  \\
&{Holiday}                            & -0.503$^{***}$ \; (0.097)  & -0.369$^{***}$ \; (0.129)  & -0.394$^{***}$ \; (0.088)  \\
\multicolumn{5}{l}{\textit{Creator Profile in QA Community}} \\
&{Follower}                           & 0.123$^{***}$ \; (0.037)   & 0.126$^{***}$ \; (0.043)   &  0.120$^{***}$ \; (0.029)  \\
&{Followee}                           & -0.001 \; (0.021)          & 0.039 \; (0.028)           &  0.003 \; (0.019)          \\
&{Answer}                             & 0.034 \; (0.024)           & 0.036 \; (0.032)           &  0.050$^{**}$ \; (0.022)   \\
&{Upvote}                             & 0.378$^{***}$ \; (0.077)   & 0.349$^{***}$ \; (0.099)   &  0.427$^{***}$ \; (0.068)  \\
&{Downvote}                           & 0.204$^{***}$ \; (0.049)   & 0.188$^{***}$ \; (0.062)   &  0.126$^{***}$ \; (0.042)  \\
&{Thank}                              & 0.004 \; (0.084)           & -0.040 \; (0.108)          & -0.166$^{**}$ \; (0.073)   \\
&{Unhelpful}                          & -0.411$^{***}$ \; (0.100)  & -0.322$^{**}$ \; (0.125)   & -0.235$^{***}$ \; (0.085)  \\
&{Badge}                              & -0.043 \; (0.063)          & -0.114 \; (0.081)          & -0.101$^{*}$ \; (0.056)    \\
&{TenureQA}                           & 0.101$^{***}$ \; (0.023)   & 0.089$^{***}$ \; (0.030)   &  0.055$^{***}$ \; (0.021)  \\
&{Celebrity}                          & 0.836$^{***}$ \; (0.148)   & 1.223$^{***}$ \; (0.192)   &  0.898$^{***}$ \; (0.132)  \\
\multicolumn{5}{l}{\textit{Creator Profile in Zhihu Live Market}} \\
&{NumPastEvent}                       & -0.152$^{***}$ \; (0.035)  & -0.180$^{***}$ \; (0.044)  & -0.139$^{***}$ \; (0.030)  \\
&{AvgPastSales}                       & 0.310$^{***}$ \; (0.043)   & 0.253$^{***}$ \; (0.055)   &  0.209$^{***}$ \; (0.037)  \\
&{AvgPastRating}                      & 0.064$^{**}$ \; (0.026)    & 0.063$^{*}$ \; (0.034)     &  0.025 \; (0.023)          \\
&{TenureMarket}                       & -0.010 \; (0.039)          & 0.024 \; (0.050)           &  0.045 \; (0.035)          \\
&{Invitation}                         & 0.067 \; (0.101)           & 0.149 \; (0.128)           &  0.089 \; (0.087)          \\
\hline
& Category fixed effects                                        & {Yes}  & {Yes}  & {Yes} \\
& Time fixed effects                                            & {Yes}  & {Yes}  & {Yes} \\
& Category $\times$ Time fixed effects                          & {Yes}  & {No}   & {No}  \\
& $Post$ $\times$ Category fixed effects                        & {No}   & {Yes}  & {No}  \\
& $Post$ $\times$ Time fixed effects                            & {Yes}  & {Yes}  & {Yes} \\
& $Post$ $\times$ Event and creator characteristics             & {No}   & {Yes}  & {No}  \\
& Event embeddings                                              & {No}   & {No}   & {Yes} \\
\hline
&R$^2$                                                          & {0.585}  & {0.501}  & {0.552} \\
&Num. obs.                                                      & {4,598}  & {4,598}  & {4,598} \\
\hline
\end{tabular}
\begin{tablenotes}[para,flushleft]
{\footnotesize Note: Each event contributes one observation for the pre-period and one for the post-period, resulting in a total of $2,299 \times 2 = 4,598$ observations. Time fixed effects refer to event start week and day-of-week fixed effects. We standardize all continuous covariates to ensure comparability of coefficients. $^{***}p<0.01$, $^{**}p<0.05$, $^*p<0.1$.}
\end{tablenotes}
\end{threeparttable}
\end{table}

\begin{table}
	\footnotesize
	\renewcommand{\arraystretch}{1}
	\addtolength{\tabcolsep}{0pt}
	\centering
	\begin{threeparttable}
		\caption{IV Estimation: Separate Pre- and Post-Period Regressions}\label{tab:iv_separate}
		\begin{tabular}{ll *{5}{S[table-format=2.3, table-space-text-post=*********]}}
		\hline
			& & \multicolumn{2}{c}{\textit{Full Controls}} & & \multicolumn{1}{c}{\textit{Exclusion of $Rating$}}\\
			\cline{3-4} \cline{6-6}
			& {Dependent Variable}   & {$Ln(Demand_{Pre})$} & {$Ln(Demand_{Post})$} & & {$Ln(Demand_{Pre})$} \\
			&  & {(1)}  & {(2)}  & & {(3)} \\
			\hline
            \multicolumn{6}{l}{\textit{Key Independent Variables}} \\
            &{\textbf{$Ln(Price)$}}      & -1.311$^{***}$ \; (0.108)   & -0.729$^{***}$ \; (0.120)   & & -1.289$^{***}$ \; (0.112) \\
			\multicolumn{6}{l}{\textit{Event Characteristics}} \\
			&{ExposureDuration}          & 0.037 \; (0.033)            & -0.041 \; (0.037)           & & 0.069$^{**}$ \; (0.035)   \\
			&{SeatLimit}                 & -0.070$^{***}$ \; (0.026)   & -0.057$^{**}$ \; (0.028)    & & -0.070$^{***}$ \; (0.027) \\
			&{Advertisement}             & 0.663$^{***}$ \; (0.107)    &  0.433$^{***}$ \; (0.118)   & & 0.697$^{***}$ \; (0.110)  \\
			&{Rating}                    & 0.442$^{***}$ \; (0.035)    &  0.496$^{***}$ \; (0.038)   & & \\
			&{Holiday}                   & -0.369$^{***}$ \; (0.122)   & -0.419$^{***}$ \; (0.135)   & & -0.348$^{***}$ \; (0.126) \\
			\multicolumn{6}{l}{\textit{Creator Profile in QA Community}} \\			
			&{Follower}                  & 0.126$^{***}$ \; (0.040)    & 0.093$^{**}$ \; (0.045)    & & 0.116$^{***}$ \; (0.042)   \\
			&{Followee}                  & 0.039 \; (0.026)            &  0.010 \; (0.029)          & & 0.041 \; (0.027) \\
			&{Answer}                    & 0.036 \; (0.030)            &  0.017 \; (0.033)          & & 0.036 \; (0.031) \\
			&{Upvote}                    & 0.349$^{***}$ \; (0.094)    &  0.443$^{***}$ \; (0.104)  & & 0.394$^{***}$ \; (0.097) \\
			&{Downvote}                  & 0.188$^{***}$ \; (0.059)    &  0.155$^{**}$ \; (0.065)   & & 0.174$^{***}$ \; (0.061) \\
			&{Thank}                     & -0.040 \; (0.102)           &  -0.064 \; (0.113)         & & -0.093 \; (0.106) \\
			&{Unhelpful}                 & -0.322$^{***}$ \; (0.119)   &  -0.360$^{***}$ \; (0.132) & & -0.294$^{**}$ \; (0.123) \\
			&{Badge}                     & -0.114 \; (0.077)           &  -0.146$^{*}$ \; (0.085)   & & -0.040 \; (0.080) \\
			&{TenureQA}                  & 0.089$^{***}$ \; (0.028)    &  0.079$^{**}$ \; (0.032)   & & 0.095$^{***}$ \; (0.029) \\
			&{Celebrity}                 & 1.223$^{***}$ \; (0.182)    &  0.333$^{*}$ \; (0.202)    & & 1.141$^{***}$ \; (0.189) \\
			\multicolumn{6}{l}{\textit{Creator Profile in Zhihu Live Market}} \\
			&{NumPastEvent}              & -0.180$^{***}$ \; (0.042)   &  -0.091$^{*}$ \; (0.046)   & & -0.217$^{***}$ \; (0.043) \\
			&{AvgPastSales}              & 0.253$^{***}$ \; (0.052)    &  0.355$^{***}$ \; (0.057)  & & 0.262$^{***}$ \; (0.054) \\
			&{AvgPastRating}             & 0.063$^{**}$ \; (0.032)     &  0.006 \; (0.035)          & & 0.103$^{***}$ \; (0.033) \\
			&{TenureMarket}              & 0.024 \; (0.048)            &  -0.030 \; (0.053)         & & 0.021 \; (0.049) \\
			&{Invitation}                & 0.149 \; (0.122)            &  0.031 \; (0.135)          & & 0.217$^{*}$ \; (0.126) \\
			\hline
            &Category fixed effects      & {Yes}                       & {Yes}                      & & {Yes} \\
			&Time fixed effects          & {Yes}                       & {Yes}                      & & {Yes} \\
			\hline 
			&R$^2$                       & {0.460}                     & {0.352}                    & & {0.421}  \\
            &Num. obs.                   & {2,299}                     & {2,299}                    & & {2,299}  \\
			\hline
		\end{tabular}
		\begin{tablenotes}[para,flushleft]
		{\footnotesize Note: Time fixed effects refer to event start week and day-of-week fixed effects. We standardize all continuous covariates to ensure comparability of coefficients. Event characteristics and creator profiles in both the Q\&A community and the Zhihu Live market are omitted for ease of tabulation. $^{***}p<0.01$, $^{**}p<0.05$, $^*p<0.1$.}
		\end{tablenotes}
	\end{threeparttable}
\end{table}

\newpage
\clearpage
\section{Mechanisms}\label{app:mechanism}

\subsection{Word-of-mouth Information}\label{app:review}

One important source of word-of-mouth information is the ratings and reviews submitted by consumers upon livestream completion. Across all events, consumers submitted a total of 77,956 ratings and 29,652 text reviews, with each event receiving an average of 35 ratings and 13 reviews. The average rating across events is 4.43 (S.D. = 1.01) and the average review length is 18.20 words (S.D. = 24.30). \Cref{tab:reviews} presents ten randomly drawn reviews from our data.
\begin{table}[!h]
    \footnotesize
    \renewcommand{\arraystretch}{1.2}
    \centering
    \begin{threeparttable}
        \caption{Examples of Reviews}
        \label{tab:reviews}
        \begin{tabular}{lp{15cm}}
            \hline
            {No.} & {Review Content} \\
            \hline
            1 & Very comprehensive and packed with useful information. \\ % 68521
            2 & The delivery isn't very fluent, and there isn't much substance to it. \\%66381
            3 & Awesome. \\ %63374
            4 & There's a lot of practical information, and it's all very concise. Maybe because I already have some background knowledge, it wasn't too hard for me to grasp. \\ %64735
            5 & The voice was a bit soft, but the presentation was professional and well-targeted; however, there were some issues with pacing. \\ %64525
            6 & The explanation was very vivid and detailed. \\ %62930
            7 & The foundational theory is decent, but I feel it didn’t quite meet my expectations. For example, the section on forgetfulness didn’t offer any effective solutions or relevant theories. \\ %61609
            8 & It was clear that the presenter put a lot of thought into it, spent a great deal of time preparing, and was well-prepared, with clear logic and precise delivery. For me, this was definitely a live session that I found extremely beneficial. \\ %57056
            9 & Great! This livestream focused more on methodology. I suggest offering more specific livestreams on different topics in the future, such as academic sessions or business sessions… Looking forward to it! \\ %50486
            10 & Takeaways: I need to dive deeper into this and put more effort into learning the foundational knowledge required… Anyway… since I want to become a tech expert… I just have to keep evolving! I thought the instructor did a great job… Applause!!! \\ %50456
            \hline
        \end{tabular}
        \begin{tablenotes}[para, flushleft]
            {\footnotesize }
        \end{tablenotes}
    \end{threeparttable}
\end{table}

% \begin{table}[!h]
%     \footnotesize
%     \renewcommand{\arraystretch}{1.2}
%     \centering
%     \begin{threeparttable}
%         \caption{Top Keywords in Latent Topics from Event Reviews}
%         \label{tab:bertopic}
%         \begin{tabular}{lp{9cm}}
%             \hline
%             {Topic Name} & {Top 10 Keywords} \\
%             \hline
%             Content Quality and Practical Value & Content, Practical, Session, Information, Helpful, Instructor, Useful, Teacher, Learned, Packed \\
%             Livestream Experience and Presentation & Live, Stream, Live stream, Zhihu, Presentation, Content, Session, Presenter, First, Listening \\
%             Overall Evaluation and Rating Expressions & Star, Thumb, Five star, Five, Review, Sister, Bad, Give, Awesome, Master \\
%             \hline
%         \end{tabular}
%         \begin{tablenotes}[para, flushleft]
%             {\footnotesize Note: Topics are ranked by the number of reviews assigned and are named based on the semantic meaning of their associated keywords. Keywords are ordered by their relevance within each topic. The top three topics together account for over 50\% of all reviews.}
%         \end{tablenotes}
%     \end{threeparttable}
% \end{table}

To assess whether reviews provide meaningful information about content quality, we conduct the following text analyses. First, we extract the most frequently occurring phrases from reviews across all events. The top-5 phrases include ``live,'' ``content,'' ``teacher,'' ``learned,'' and ``practical,'' suggesting a strong focus on learning and content quality. Second, we apply BERTopic \citep{grootendorst2022bertopic}, a transformer-based topic modeling technique that extracts semantically coherent topics from text. \Cref{tab:bertopic} in Section \ref{sec:mechanism} presents the keywords associated with the top-3 topics, which consistently center on content quality, information, and learning. Taken together, these findings support that consumer reviews provide useful quality-related information for prospective buyers in the post-period.

% ZC: Please feel free to drop this paragraph if you find the LIWC analysis provide little insights
As a supplementary check, we examine review linguistic styles using Linguistic Inquiry and Word Count (LIWC) \citep{boyd2022development}. Reviews in our sample score relatively high on the Authenticity dimension (mean = 61.8, on a scale of 0 to 99), suggesting that the reviews are generally expressed in a personal and natural manner.

% \begin{table}[!h]
%     \footnotesize
%     \renewcommand{\arraystretch}{1.2}
%     \centering
%     \begin{threeparttable}
%         \caption{LIWC Analysis of Event Reviews}
%         \label{tab:liwc}
%         \begin{tabular}{lp{8cm}rrr}
%             \hline
%             {Variable} & {Definition} & {Mean} & {Std. Dev.} & {Theoretical Range} \\
%             \hline
%             WC        & Total word count                                                       & 18.20 & 24.30 & $0 - \infty$ \\
%             Analytic  & Analytical thinking (high = formal/logical, low = narrative/emotional) & 38.56 & 35.88 & $1 - 99$ \\
%             Clout     & Social influence / confidence level                                    & 32.80 & 38.01 & $1 - 99$ \\
%             Authentic & Authenticity / candor                                                  & 61.78 & 36.91 & $1 - 99$ \\
%             Tone      & Emotional tone (high = positive, low = negative)                       & 84.28 & 29.62 & $1 - 99$ \\
%             \hline
%         \end{tabular}
%         \begin{tablenotes}[para, flushleft]
%             {\footnotesize Note: Analytic, Clout, Authentic, and Tone are composite indices ranging from 1 to 99. The high Tone score (84.3) indicates that reviews are predominantly positive in emotional tone, and the high Authentic score (61.8) suggests that reviews are written in a candid and personal style.}
%         \end{tablenotes}
%     \end{threeparttable}
% \end{table}

\subsection{Heterogeneity across Creators and Content Category}
\begin{table}[!h]
\small
\renewcommand{\arraystretch}{1}
\addtolength{\tabcolsep}{0pt}
\centering
\begin{threeparttable}
\caption{IV Estimation: Heterogeneity by Consumer Follow Status (Intersection Group)}\label{tab:cons_seg_iv_intersection}
\begin{tabular}{ll *{2}{S[table-format=2.3, table-space-text-post=***]}}
\hline
&  {Dependent Variable} & {$Ln(Demand^{NF}$)} & {$Ln(Demand^{F}$)} \\
& & {(1)} & {(2)}  \\
\hline
&{$Ln(Price)$}             & -1.163$^{***}$ & -0.626$^{***}$  \\
&                          & {(0.104)}      & {((0.098))}     \\
&{$Post$}                  & -2.367$^{***}$ & -2.305$^{***}$  \\
&                          & {(0.848)}      & {(0.795)}       \\
&{$Post \times Ln(Price)$} & 0.421$^{***}$  & 0.337$^{***}$  \\
&                          & {(0.136)}      & {(0.127)}     \\
&Event characteristics     & {Yes}          & {Yes}           \\
&Creator characteristics   & {Yes}          & {Yes}           \\
\hline
&Category fixed effects    & {Yes}          & {Yes}           \\
&Time fixed effects        & {Yes}          & {Yes}           \\
&$Post \times$ time fixed effects  & {Yes}  & {Yes}   \\
\hline
&R$^2$                     & {0.488}        & {0.441}        \\
&Num. obs.                 & {4,598}        & {4,598}        \\
\hline
\end{tabular}
\begin{tablenotes}[para,flushleft]
{\footnotesize Note: Each event contributes one observation for the pre-period and one for the post-period, resulting in a total of $2,299 \times 2 = 4,598$ observations. Time fixed effects refer to event start week and day-of-week fixed effects. We standardize all continuous covariates to ensure comparability of coefficients. Event characteristics and creator profiles in both the Q\&A community and the Zhihu Live market are omitted for ease of tabulation. $^{***}p<0.01$, $^{**}p<0.05$, $^*p<0.1$.}
\end{tablenotes}
\end{threeparttable}
\end{table}

\begin{sidewaystable}
\footnotesize
\renewcommand{\arraystretch}{1}
\addtolength{\tabcolsep}{0pt}
\centering
\begin{threeparttable}
\caption{IV Estimation: Heterogeneity by Creator Reputation and Content Category (Intersection Group)}\label{tab:event_seg_iv_intersection}
\begin{tabular}{ll *{10}{S[table-format=2.3, table-space-text-post=***]}}
\hline
& & \multicolumn{10}{c}{Dependent Variable: $Ln(Demand)$} \\
\cline{3-12}
&  {Event Segments} & {Non-recipient} & {Recipient} & & {$F^1$} & {$F^2$} & {$F^3$} & {$F^4$} & {$F^5$} & {Soft} & {Hard} \\
& & {(1)} & {(2)} & & {(3)} & {(4)} & {(5)} & {(6)} & {(7)} & {(8)} & {(9)}  \\
\hline
& {$Ln(Price)$}            & -1.101$^{***}$ & -1.012$^{***}$  &  & -1.188$^{***}$ & -1.372$^{***}$   & -0.483$^{**}$  & -1.371$^{***}$ & -1.363$^{***}$ & -1.180$^{***}$ & -1.000$^{***}$ \\
&                          & {(0.128)}      & {(0.181)}       &  & {(0.347)}      & {(0.291)}        & {(0.218)}      & {(0.225)}      & {(0.165)}      & {(0.154)}      & {(0.141)}      \\
&{$Post$}                  & -2.585         & -2.513$^{***}$  &  & -3.219$^{***}$ & -2.422$^{**}$    & -2.802$^{***}$ & -2.057$^{**}$  & -3.341$^{***}$ & -2.693         & -2.274$^{**}$  \\
&                          & {(1.668)}      & {(0.830)}       &  & {(1.205)}      & {(1.113)}        & {(1.135)}      & {(1.038)}      & {(1.067)}      & {(1.646)}      & {(0.939)}      \\
&{$Post \times Ln(Price)$} & 0.537$^{***}$  & 0.376$^{**}$    &  & 0.749          & 0.842$^{**}$     & 0.467$^{*}$    & 0.385          & 0.469$^{**}$   & 0.621$^{***}$  & 0.271          \\
&                          & {(0.173)}      & {(0.188)}       &  & {(0.471)}      & {(0.399)}        & {(0.278)}      & {(0.297)}      & {(0.187)}      & {(0.200)}      & {(0.177)}      \\
&Event characteristics     & {Yes}          & {Yes}           &  & {Yes}          & {Yes}            & {Yes}          & {Yes}          & {Yes}          & {Yes}          & {Yes}          \\
&Creator characteristics   & {Yes}          & {Yes}           &  & {Yes}          & {Yes}            & {Yes}          & {Yes}          & {Yes}          & {Yes}          & {Yes}          \\
\hline
&Category fixed effects    & {Yes}          & {Yes}           &  & {Yes}          & {Yes}            & {Yes}          & {Yes}          & {Yes}          & {Yes}          & {Yes}          \\
&Time fixed effects        & {Yes}          & {Yes}           &  & {Yes}          & {Yes}            & {Yes}          & {Yes}          & {Yes}          & {Yes}          & {Yes}          \\
&$Post \times$ time fixed effects  & {Yes}  & {Yes}           &  & {Yes}          & {Yes}            & {Yes}          & {Yes}          & {Yes}          & {Yes}          & {Yes}          \\
\hline
&R$^2$                     &  {0.506}       &  {0.664}        &   & {0.656}       & {0.570}          & {0.557}        & {0.579}        & {0.669}        & {0.510}        & {0.554}        \\
&Num. obs.                 &  {3,778}       &  {820}          &   & {920}         & {920}            & {920}          & {919}          & {919}          & {2,498}        & {2,100}        \\
\hline
\end{tabular}
\begin{tablenotes}[para,flushleft]
{\footnotesize Note: $F^{g}$ ($g \in \{1, 2, 3, 4, 5\}$) is the segment of events within the $g$th quintile range of $Follower_{j_{it}}$. Within each segment, every event contributes one observation for the pre-period and one for the post-period. Time fixed effects refer to event streaming week and day-of-week fixed effects. We standardize all continuous covariates to ensure comparability of coefficients. Event characteristics and creator profiles in both the Q\&A community and the Zhihu Live market are omitted for ease of tabulation. $^{***}p<0.01$, $^{**}p<0.05$, $^*p<0.1$.}
\end{tablenotes}
\end{threeparttable}
\end{sidewaystable}

\clearpage
\newpage
\begin{figure}[!h]
	\centering
	\includegraphics[width=1\hsize]{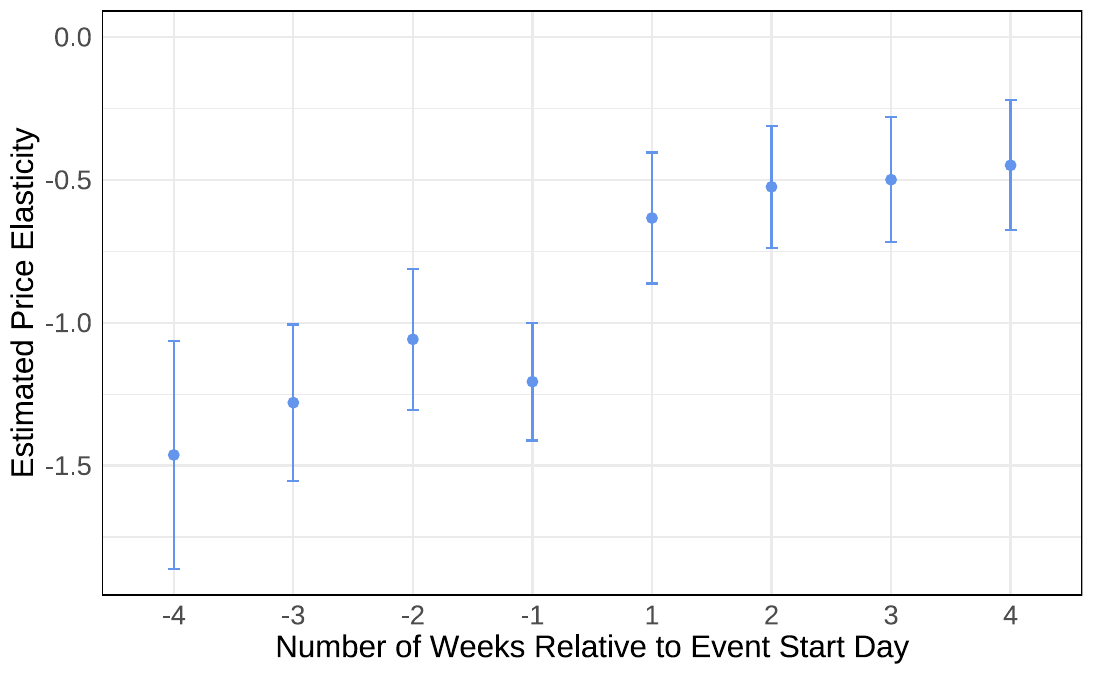}
	\caption{Price Elasticity Over Number of Weeks to Event Start Day}\label{fig:ranking_effect}
	\begin{minipage}{0.9\textwidth} % choose width suitably
		{\footnotesize Note: Week –1 (1) refers to the seven-day period immediately before (after) the event start date, inclusive of the start day. Week –2 (2) denotes the seven days preceding (following) Week –1 (1), and so on.}
	\end{minipage}
\end{figure}

\end{document}